\documentclass[letterpaper]{article} 
\usepackage{aaai25}  
\usepackage{times}  
\usepackage{helvet}  
\usepackage{courier}  
\usepackage[hyphens]{url}  
\usepackage{graphicx} 
\usepackage{natbib}  
\usepackage{caption} 
\usepackage{algorithm}
\usepackage{algorithmic}
\usepackage{tikz}
\usepackage{cite}
\usepackage{amsmath,amssymb,amsfonts}
\usepackage{amsthm}
\usepackage{algorithmic}
\usepackage{textcomp}
\usepackage{xcolor}
\newtheorem{theorem}{Theorem}

\newtheorem{lemma}{Lemma}
\newtheorem{assumption}{Assumption}
\newtheorem{proposition}{Proposition}
\newtheorem{remark}{Remark}
\usepackage{multirow}
\usepackage{hhline}
\usepackage{threeparttable}
\usepackage{tabularray}
\usetikzlibrary{positioning} 
\usepackage{newfloat}
\usepackage{listings}
\DeclareCaptionStyle{ruled}{labelfont=normalfont,labelsep=colon,strut=off} 
\floatstyle{ruled}
\newfloat{listing}{tb}{lst}{}
\floatname{listing}{Listing}
\title{Single-Loop Federated Actor-Critic across Heterogeneous Environments}
\author{
Ye Zhu\textsuperscript{\rm 1}, Xiaowen Gong\textsuperscript{\rm 1}
}
\affiliations{
    \textsuperscript{\rm 1}Auburn University, Auburn, AL, USA\\

\{yzz0211,xzg0017\}@auburn.edu%
}

\usepackage{bibentry}

\begin{document}

\maketitle

\begin{abstract}
Federated reinforcement learning (FRL) has emerged as a promising paradigm, enabling multiple agents to collaborate and learn a shared policy adaptable across heterogeneous environments. Among the various reinforcement learning (RL) algorithms, the actor-critic (AC) algorithm stands out for its low variance and high sample efficiency. However, little to nothing is known theoretically about AC in a federated manner, especially each agent interacts with a potentially different environment. The lack of such results is attributed to various technical challenges: a two-level structure illustrating the coupling effect between the actor and the critic, heterogeneous environments, Markovian sampling and multiple local updates. In response, we study \textit{Single-loop Federated Actor Critic} (SFAC) where agents perform actor-critic learning in a two-level federated manner while interacting with heterogeneous environments. We then provide bounds on the convergence error of SFAC. The results show that the convergence error asymptotically converges to a near-stationary point, with the extent proportional to environment heterogeneity. Moreover, the sample complexity exhibits a linear speed-up through the federation of agents. We evaluate the performance of SFAC through numerical experiments using common RL benchmarks, which demonstrate its effectiveness.
\end{abstract}

\section{Introduction}

In single-agent reinforcement learning (RL), policy improvement can be effectively facilitated by the Policy Gradient Theorem \cite{sutton99} when using parameterized policies. This theorem represents the policy gradient as a product of the score function and the action-value function, serving as the cornerstone for numerous RL algorithms. Among these, actor-critic (AC) algorithms stand out by employing temporal difference (TD) learning to approximate the action-value function, unlike Monte-Carlo algorithms \cite{williams}, which rely on sampling entire trajectories to estimate the value function. AC algorithms not only make better use of available samples but also significantly reduce the variance in policy updates, leading to more stable and efficient learning.


In AC algorithms, there are two main methods: double-loop variants \cite{kumar2023sample, improving_liang} and single-loop variants \cite{qiu,alset, ttsa, stac}. In non-asymptotic analysis of AC, the double-loop variants facilitate a decoupled convergence analysis of the critic and the actor, involving a policy evaluation sub-problem in the inner loop and a perturbed gradient descent in the outer loop. While the double-loop variant simplifies the convergence analysis, it is rarely adopted in practice due to its requirement for accurate critic estimation \cite{kumar2023sample}, making it typically sample-inefficient. The key distinction between double-loop and single-loop variants lies in their methods for achieving convergence. Double-loop AC requires a sufficient number of iterations for value evaluation in the inner loop to ensure convergence. In contrast, the single-loop method eliminates this requirement by preserving the critic’s memory throughout the process, providing a significant advantage in terms of efficiency.

\vspace{-0.3em}
Federated reinforcement learning (FRL) is an emerging distributed learning framework that integrates the key concepts of federated learning (FL) and RL \cite{fed_fault,fedtd,fedkl,ciss}. Its growing prominence is driven by its versatility in real-world applications such as autonomous driving \cite{auto_driving_1,auto_driving_2} and resource allocation \cite{resource1,resource2}. Existing FRL methods, such as FedSARSA \cite{fedsarsa} and FedTD \cite{fedtd,fedhertd}, are widely applied to stochastic problems with single-level structure. Instead, FRL with nested formulations, has not been studied. In this paper, we propose single-loop federated actor critic (SFAC), where agents in heterogeneous environments perform actor-critic learning in a two-level federated manner. Specifically, SFAC is composed of federated critics (FedC) where critics perform heterogeneous federated TD learning for policy evaluation by aggregating their local value function models, and federated actors (FedA) where the actors perform federated policy update for policy improvement by aggregating their local policy gradients. Through local information aggregation, SFAC enables the learning of a shared policy that achieves strong average performance across all agents' heterogeneous environments.
\begin{table*}[t]
\captionsetup{skip=2pt}
\label{comparison-table}
\centering
\begin{threeparttable}
\begin{tabular}{c c c c c c}
\hline
\multirow{2}{*}{References} &Number of & Heterogeneous  & Target  & Markovian & Linear \\ &Agents &Environments & Environment&Sampling&Speedup\\
\hline
\cite{alset} &1& $-$ & $-$ & $ \times $ & $-$ \\
\hline
\cite{stac} &1 & $-$ & $-$ & $ \surd $ & $-$ \\
\hline
\cite{a3c}  & $N$ & $ \times $ & Individual & $\surd $~ & IID setting \\
\hline
This paper &$N$& $\surd $~ & Mixture~  & $\surd $~ & $\surd $\\     
\hline
\end{tabular}
\caption{Comparison of Settings and Results with the Most Related Works: "Target environment" is the benchmark for evaluating FRL; "Individual" means the target environment is the same for all agents since all agents interact with (different instances of) the same environment; "IID setting" means the linear speedup result is only established for the IID setting.}

\end{threeparttable}
\setlength{\abovecaptionskip}{0.2cm}
\setlength{\belowcaptionskip}{-0.2cm}
\end{table*} 
In addition to common challenges in FRL, such as Markovian sampling and multiple local iterations, the convergence analysis of SFAC encounters several unique difficulties stemming from its distinctive features. First, two levels of local model aggregations among agents' critics and actors have coupled and non-trivial impacts on the convergence error of SFAC. Second, single-loop structure brings the analysis of SFAC substantial errors in critic estimation and the tight coupling between parallel updates of critics and actors, making the algorithm more susceptible to unstable error propagation. Third, due to the environment heterogeneity, agents in SFAC have distinct optimal value functions for a given policy and also different optimal policies, causing non-trivial biases in the convergence analysis. Lastly, to measure the performance of the common policy obtained by SFAC for all agents' environments, we focus on the \textit{mixture} environment which is \textit{randomly drawn} from agents' heterogeneous environments. Compared to the virtual environment considered in recent works on FRL~\cite{fedhertd,fedsarsa,aistat22}  (which is defined by directly averaging transition probability kernels and reward functions of agents' environments), the mixture environment presents a more meaningful yet significantly more complex challenge for convergence analysis.

The main contributions of this paper are summarized as follows:
\begin{itemize}
\item \textbf{SFAC algorithm.} We devise \textit{Single-loop Federated Actor Critic} (SFAC) where agents perform actor-critic learning across heterogeneous environments in a two-level federated manner. The two-level structure introduces a non-trivial coupling between parallel updates of critics and actors. 

\item \textbf{SFAC analysis.} 
We develop a new analysis framework that establishes the finite-time convergence for SFAC. Technically, we exploit the property of the global gradient of the average mean-squared projected Bellman error (MSPBE) in the critic's error, so that a biased term in the critics can be eliminated in the actors. The results show that the convergence error is asymptotically determined by the heterogeneity of agents' environments, and it diminishes to 0 as the environments heterogeneity reduces to 0. Moreover, the sample complexity enjoys linear speed-up through federation. 
\end{itemize}

\section{Related Works}
\noindent\textbf{Actor-Critic algorithms.}
Most of the existing theoretical work on AC focuses on the single-agent case. \cite{kumar2023sample,PGgloptimal,qiu,nips20,improving_liang}. Even within multi-agent AC frameworks, the problem is frequently approached as if it were a single-agent AC issue, viewed from a global perspective. For double-loop AC, the non-asymptotic analyses have been well established in both IID sampling and Markovian sampling. For single-loop AC, through the lens of bi-level optimization, two-timescale AC \cite{ttsa} and single-timescale AC \cite{stac, alset} have been proposed, achieving sample complexities of $\tilde O({\epsilon ^{ - 2.5}})$ and $\tilde O({\epsilon ^{ - 2}})$ respectively. The most closely related work to our paper that multiple agents executes in independent environments and collectively seeks a global policy using AC algorithms is \cite{a3c}. However, this algorithm is executed in homogeneous environments and the architecture is the shared memory that is accessible to all agents. Besides, \cite{a3c} only established the linear speedup result under IID sampling. [see Table 1 for details].

\noindent\textbf{Federated bilevel optimization.}
Since AC algorithm with linear value function approximation can be viewed as a special case of bilevel algorithms \cite{mengdi, alset, ttsa}, it is worth noting that SFAC is not a special case of federated bilevel learning (FBO). In FBO \cite{fednest, fedmbo, simfbo}, the optimization problem is
\begin{subequations}
\begin{equation}
\mathop {\min }\limits_{x \in {{\mathbb{R}}^{{d_1}}}} f(x,{y^*}(x)): = \frac{1}{m}\sum\limits_{i = 1}^m {{f_i}(x,{y^*}(x))}   
\end{equation}
\begin{equation}
s.t.\;\;{y^*}(x) = \mathop {{\rm{argmin}}}\limits_{y \in {{\mathbb{R}}^{{d_2}}}} {\rm{ }}g(x,y): = \frac{1}{m}\sum\limits_{i = 1}^m {{g_i}(x,y)}
\end{equation}
\end{subequations}
Note that the upper-level optimization in FBO is performed on ${f_i}(x,y^*(x))$ while the upper-level optimization in FAC is performed on ${f_i}(x,y_i^*(x))$. Furthermore, $g(x,y)$ has a special definition in FRL and needs to be addressed carefully.

\noindent\textbf{Federated reinforcement learning.}
FRL differs from distributed reinforcement learning \cite{zhang2018fully,zhang2021decentralized} in several key ways: 1) agents in FRL interact with heterogeneous environments and follow their respective MDPs; 2) the architecture of agents in FRL is worker-server, with a central server that coordinates with $N$ agents; 3) FRL involves some unique features of FL, including multiple local iterations of agents, heterogeneous and time-varying computation capabilities of agents.

There have been a few recent works on FRL~\cite{fedsarsa,fedhertd,aistat22}. \cite{fedhertd} considered the federated policy evaluation; \cite{fedsarsa, aistat22} analyzed the federated action value iteration. However, none of them has established the convergence of FRL algorithms by considering the collaboration of policy evaluation and policy improvement simultaneously as a two-level collaboration. Furthermore, these three papers aims to solve optimization problem in a constructed virtual environment. Thus, these works for the federation of (action) value functions do not apply to our setting. Specifically, it remains unclear how the federation of critics affect the convergence performance of the optimal policy for the mixture environment considered in this paper. In this paper, we show that the proposed algorithm can asymptotically produce a near-stationary point of the mixture environment, which is the \textit{first} result in existing works on FRL. 
\section{Preliminaries}
\subsection{MDP in Heterogeneous Environments}
The discounted Markov decision process (MDP) for agent $i$ is defined as ${{ M}^i} \buildrel \Delta \over = \{ { S},{ A},{{ P}_i},{{ R}_i},{\gamma}\}$, where $S$ denotes the set of states, $A$ represents the set of actions, $P_i$ is the transition kernel at agent $i$, $R_i$ is the reward function at agent $i$, and $\gamma$ is the discount factor. In this paper, while all agents share the same state-action space, their reward functions and transition kernels may vary. Specifically, when agent $i$ takes an action $a \in A$ in state $s \in S$, it transitions to the next state $s' \in S$ based on the probability distribution ${P_i}(s'\left| {s,a} \right.)$ and receives a reward ${r_i}(s,a,s')$. We consider a stochastic policy $\pi$ that assigns a probability distribution $D ( \cdot \left| s \right.)$ over the entire action space $A$ for each state $s \in S$.

Given a policy $\pi$, the state value function and the state-action value function for agent $i$ are defined as follows:
\begin{align*}
V_\pi^i (s) &= \mathbb{E}\left[ {\sum\limits_{t = 0}^\infty  {{\gamma ^t}{r_i}(s_t^i,a_t^i,s_{t+1}^i)\left| {s_0^i = s, \pi, {P_i}} \right.} } \right]\\
Q_\pi^i (s,a) &= \mathbb{E}\left[ {\sum\limits_{t = 0}^\infty  {{\gamma ^t}{r_i}(s_t^i,a_t^i,s_{t+1}^i)\left| {s_0^i = s, a_0^i = a, \pi, {P_i}} \right.} } \right] 
\end{align*}
where the expectation $\mathbb{E}$ is taken over all possible trajectories of agent $i$, given policy $\pi$. Let $b_i$ represent the initial state distribution of agent $i$. The discounted state visitation measure induced by policy $\pi$ is then defined as $\nu^i_\pi (s) = (1 - \gamma )\sum\nolimits_{t = 0}^\infty  {{\gamma ^t}{P_i}(s_t^i = s\left| {s_0^i \sim {b_i},\pi ,} \right.{P_i})}$ and the state-action visitation distribution is defined as $\nu_\pi^i (s,a) = (1 - \gamma )\sum\nolimits_{t = 0}^\infty  {{\gamma ^t}{P_i}(s_t^i = s,a_t^i = a\left| {s_0^i \sim {b_i},\pi ,} \right.{P_i})} $. 

The performance of policy $\pi$ at agent $i$ can be evaluated by the expected cumulative rewards, given by  ${J_i}(\pi ) = (1 - \gamma ){\mathbb{E}_{s \sim {\eta_i}}}\left[ {V_\pi^i (s)} \right]$. In the context of single-agent reinforcement learning, the objective is to find an optimal policy that maximizes ${J_i}(\pi)$.

\subsection{Optimal Policy for Mixture Environment}
In this paper, we tackle a federated reinforcement learning problem involving $N$ agents who collaboratively work together to find a globally optimal policy without sharing their raw collected samples. Each agent operates within its own unique environment, with each environment defined by its own MDP. Previous research on FRL \cite{aistat22, fedhertd, fedsarsa} has focused on optimizing the value function model for a single virtual environment. This virtual environment is constructed as an MDP ${ {\bar M}} \buildrel \Delta \over = \{ { S},{ A},{ {\bar P}},{ {\bar R}},{\gamma}\}$ by directly averaging the transition kernels and reward functions across all agents' environments, given by $\bar P = \frac{1}{N}\sum\nolimits_{i = 1}^N {{P_i}} $ and  $\bar R = \frac{1}{N}\sum\nolimits_{i = 1}^N {{R_i}} $. 

 However, such an "averaged" environment may not correspond to any agent's specific environment. Consequently, from the perspective of an individual agent, the objective function may not effectively incentivize participation in the federation. Motivated by this observation, we propose considering a \textit{mixture} environment in this paper, defined as an environment \textit{randomly drawn} from the set of agents' heterogeneous environments. This mixture environment differs from the virtual environment described in \cite{aistat22, fedhertd, fedsarsa}. Specifically, the virtual environment defines an MDP where, after an agent selects an environment and conducts a transition, transitioning to the successive state still demands a new selection of an environment. In contrast, in the mixture environment, once an environment is chosen with a certain probability, the state transitions follow the corresponding MDP of that environment. Compared to the virtual environment, the mixture environment is more practical. 
 
 Therefore, the goal of agents in FRL is to cooperatively find an optimal policy ${\omega ^*}$ that maximizes the total cumulative discounted rewards.
\begin{align}
{\theta ^*} = \arg \mathop {\max }\limits_\theta  J(\theta ) = \arg \mathop {\max }\limits_\theta  \frac{1}{N}\sum\limits_{i = 1}^N {{J_i}(\theta )},
\end{align}
where ${{J_i}(\theta )}$ is a non-concave objective function. We use ${\pi _\theta }$ to denote the policy parameterized by $\theta  \in {{\mathbb{R}}^d}$. With no ambiguity, we use $\theta$ to denote the policy ${\pi _\theta }$. 

Treating each environment as an independent RL problem implies that each agent aims to maximize their own ${{J_i}(\theta )}$. However, our goal is to develop a single policy that optimally balances performance across all environments, while preserving the unique characteristics of each agent's environment. This objective aligns with the principles of federated supervised learning.

\subsection{Actor-Critic with Function Approximation}
In this subsection, we will provide a detailed introduction to AC with function approximation.
\\
\textbf{Critic update.}
In AC, (critics do) policy evaluation and (actors do) policy improvement are alternately applied to achieve the optimal policy. A common approach in policy evaluation is to use linear function approximation. Let $\{ {\Phi _k}\} _{k = 1}^d$ be a set of $d$ linearly independent basis vectors in ${{\mathbb{R}}^n}$. The true value function ${V_\pi }$ is then approximated as ${V_\pi }(s) \approx {V_\omega }(s) = \phi {(s)^{\rm{T}}}\omega$, where $\phi (s) \in {{\mathbb{R}}^d}$ is a fixed feature vector for state $s$ and $\omega \in {{\mathbb{R}}^d}$ is the unknown model to be learned. For convenience, we drop the subscript (i) since each agent follows TD-learning to update its local value function. For an observed tuple $({s},{r},{s'})$, the TD operator is defined as follows, for any $V \in {{\mathbb{R}^{|S|}}}$,
\begin{equation} \label{td target}
({T_\pi }V)(s) = {R_\pi }(s) + \gamma \sum\limits_{s'} {{P_\pi }} (s,s')V(s'). 
\end{equation}
Then the value function satisfies the Bellman equation ${T_\pi }{V_\pi } = {V_\pi }$.

The task of finding $\omega^*$ is usually addressed by TD learning. The loss function is defined as the squared Bellman error \cite{PMLR2018}. At time $t$, an agent observes samples ${O_t} = ({s_t},{r_t},{s_{t + 1}})$ and the negative stochastic semi-gradient of the Bellman error evaluated at the current parameter $\omega_t$ is expressed as
\begin{equation}\label{local gradient}
\hat g({\omega _t}) = ({r_t} + \gamma \phi {({s_{t + 1}})^{\rm{T}}}{\omega _t} - \phi {({s_t})^{\rm{T}}}{\omega _t})\phi ({s_t}).
\end{equation}
Then the estimated model at time $k+1$ can be updated by the gradient descent method \cite{PMLR2018} with step size $\beta$ as 
\begin{equation}\label{td iteration}
{\omega _{t + 1}} = {\omega _t} + \beta \hat g({\omega _t})
\end{equation}
\textbf{Actor update.} 
Policy improvement using policy function approximation is typically performed through the policy gradient theorem. For agent $i$, the gradient $\nabla {J_i}(\theta )$ is derived by the policy gradient method \cite{sutton99} as follows:
\begin{equation*}
\nabla {J_i}(\theta ) = {\mathbb{E}_{v_{\theta}^i}}\left[ {Q_{\theta}^i(s,a){\psi_ {\theta}}(s,a)} \right] = {\mathbb{E}_{v_{\theta}^i}}\left[ {A_{\theta}^i(s,a){\psi _{\theta}}(s,a)} \right]    
\end{equation*}
where ${\psi _{\theta}}(s,a) = {\nabla _{\theta}}\log {\pi _{\theta}}(a\left| s \right.)$ denotes the score function and ${v_{\theta}^i}$ denotes the state visitation measure under policy ${\pi _{\theta}}$ at agent $i$. Advantage function ${A_{{\theta }}^i(s,a)}$ is approximated by temporal difference error $\delta _{\omega_{\theta}} ^i(s,a,s') = {r_i}(s,a,s') + \gamma V_{\omega_{\theta}} ^i(s') - V_{\omega_{\theta}} ^i(s)$ where $V_{\omega_{\theta}} ^i$ is agent $i$'s critic estimate for policy ${\pi _{\theta}}$. The expected policy gradient at agent $i$ using a mini-batch can then be estimated as  
\begin{align*}
{{\hat h}_k^i}({\omega _\theta },{\theta}) = \frac{1}{M}\sum\limits_{m = 1}^M {\delta _{{\omega _\theta }}^i(s_{k,m}^i,a_{k,m}^i,s_{k,m + 1}^i){\psi _{\theta}}(s_{k,m}^i,a_{k,m}^i)} 
\end{align*}
Then agent $i$ updates its local policy by the gradient ascent method with step size $\alpha$ as
\begin{align}\label{updatepolicy}
\theta_{k + 1}^i = \theta_k^i + \alpha {{\hat h}_k^i}({\omega _\theta },\theta)
\end{align}

\section{Single-Loop Federated Actor-Critic}
In this section, we describe the single-loop federated reinforcement learning method in heterogeneous environments where agents collectively seek to find a global optimal policy. 


First, we design a single-loop process for FAC as shown in Algorithm 1. Namely, the outer loop consists of actor's updates for the policy $\pi$ to optimize the global policy, followed by an entire inner loop of critics' updates.
\\
\textbf{Inner optimizer: FedC.} We first define the global optimal value function in the mixture environment. In the process of policy evaluation for a single agent $i$, the local loss function $F_i$ is usually defined as expected Bellman error squared \cite{PMLR2018,srikant19colt}. Accordingly, the optimization problem for federated value evaluation can be formulated as
\begin{equation}\label{objective}
\mathop {\min }\limits_{\omega  \in {{\mathbb{R}}^d}} \left[ {F(\omega ) = \frac{1}{N}\sum\limits_{i = 1}^N {{F_i}(\omega )} } \right]  
\end{equation}
where ${F_i}(\omega ) = {{\mathbb{E}}_{{O_k} \sim {D _i}}}\left[ {\frac{1}{2}{{({r_k} + \gamma {V_\omega }({s_{k + 1}}) - {V_\omega }({s_k}))}^2}} \right]$ is the local objective function of $i$-th agent, i.e., the expected squared Bellman error with respect to the model parameter $\omega$ and ${{D_i}}$ is the stationary distribution of the associated state transition Markov chain in $i$-th environment.

The global objective function of our federated TD learning problem is the \textit{average MSPBE} of all agents for their respective environments, as the MSPBE quantifies the error of a value function model for an environment.  
To minimize the average MSPBE, we devise the FedC algorithm which updates the value function model via federated TD learning. The algorithm aims to iteratively estimate the gradient of the global objective function (i.e., the average MSPBE), given by
\begin{equation}\label{global_grad}
 g(\omega ) = \frac{1}{N}\sum\limits_{i = 1}^N {{{ g}_i}(\omega )}     
\end{equation}
where ${{{ g}_i}(\omega )}$ is the expected gradient of agent $i$'s local objective function (i.e., the MSPBE for agent $i$'s environment). The optimal value function model $\omega^*$ that minimizes the average MSPBE of agents satisfies $  g(\omega^*)=0$. Note that the gradient in TD learning is different from that of the standard gradient descent, as ${{{ g}_i}(\omega )}$ or ${{{ g}}(\omega )}$ is not the gradient of any \textit{fixed} objective function.

In FedC, in communication round $t$ of critics, each agent performs $\upsilon_i$ local iterations to approximate the value function of the given policy using its local observations.  $\upsilon_i$ may vary across agents since agents have \textit{heterogeneous} computation capabilities. Then agent $i$ sends the local value model $\omega_{t,{\upsilon_i}}^i$ to the global critic server. In round $t+1$, the global critic server aggregates local critics' models as 
\begin{equation}\label{globaltdmodel}
{\theta _{t + 1}} = {\Pi _{2,\cal H}}\left( {{\theta _t} + \alpha \left( {\frac{1}{N}\sum\limits_{i = 1}^N {\upsilon_i} } \right) \cdot \frac{1}{N}\sum\limits_{i = 1}^N {d_t^i} } \right)
\end{equation}
where ${\upsilon_i}$ is the number of local updates at agent $i$ and $d_t^i$ is the normalized gradient for agent $i$ in the $t$-th round as $ d_t^i = \frac{1}{{{\upsilon_i}}}\sum\limits_{k = 0}^{{\upsilon_i} - 1} {{\hat g_i}(\omega _{t,\upsilon}^i)} 
$. Here we consider agents have heterogeneous number of local updates while the number of local updates are identical and fixed in the previous work \cite{fedhertd,fedsarsa,aistat22}. Note that the cumulative local gradients are normalized by averaging, and this is necessary when dealing with heterogeneous number of local updates. Besides, we use ${\Pi _{2,\cal H}}\left(  \cdot  \right)$ to denote the standard Euclidean projection on to a convex compact subset ${\cal H }\subset {{{\mathbb{R}}^d}}$ that is assumed to contain $\omega^*$. Such a projection step is commonly adopted in RL \cite{PMLR2018, TTD}.

\begin{algorithm}[!h]\label{FedC}
    \caption{Federated Critic (FedC)}
    \label{alg:AOA}
    \renewcommand{\algorithmicrequire}{\textbf{Input:}}
    \renewcommand{\algorithmicensure}{\textbf{Output:}}
    \begin{algorithmic}[1]
        \STATE \textbf{Input}: $\omega_{k,t}$, stepsize $\beta_k$
        \STATE $\omega _{k,t}^i = \omega _{k,t}$ for agent $i \in \cal{N}$ 
        \FOR {$i \in \cal{N}$ \textbf{in parallel}}
            \FOR{$\upsilon  = 0$ to $\upsilon_i - 1$}
                \STATE Observe a tuple $O_{k,t,\upsilon}^i = (s_{k,t,\upsilon}^i,r_{k,t,\upsilon}^i,s_{k,t,\upsilon+1}^i)$ and calculate the gradient by (\ref{local gradient})
                \STATE Update the local model $\omega _{k,t,\upsilon  + 1}^i = \omega _{k,t,\upsilon }^i + {\beta _i}{{\hat g}_i}(\omega _{k,t,\upsilon }^i)$
            \ENDFOR
        \ENDFOR
        \STATE Server computes the global model by (\ref{globaltdmodel})
        \STATE \textbf{Output}: ${\omega _{k,t + 1}}$
    \end{algorithmic}
\end{algorithm}
\noindent
\textbf{Outer optimizer: FedA.} For Markovian sampling, we maintain separate Markov chains for actors and critics. For critics, samples are generated following the transition kernel $P_i$, while the actor's chain can be viewed as evolving under transition kernel $\hat{P_i}=\gamma {P_i}+(1-\gamma)\eta_i$. This separate sampling protocol for the actor and critic is essential; otherwise, using the same samples for both will result in a non-diminishing bias \cite{a3c}. The actor server aggregates local actors' policies as
\begin{equation}\label{globalpolicy}
{\theta_{k + 1}} = {\theta_k} + \alpha_k  \cdot \frac{1}{N}\sum\limits_{i = 1}^N {{{\hat h}_k^i}({{\omega }_k},{\theta_k})} 
\end{equation}
where ${{\omega }_k} = {{\omega }_{k, {T}}}$; $T$ is the number of communication rounds for FedC when evaluating the policy $\pi_{\theta_k}$. 
\begin{algorithm}[!h]\label{FedA}
    \caption{Federated Actor (FedA)}
    \label{alg:AOA}
    \renewcommand{\algorithmicrequire}{\textbf{Input:}}
    \renewcommand{\algorithmicensure}{\textbf{Output:}}
    \begin{algorithmic}[1]
        \STATE \textbf{Input}: ${\omega _{k,T}}$, stepsize $\alpha_k$, 
        \FOR {$i \in \cal{N}$ \textbf{in parallel}}
            \FOR{$m  = 0$ to $M-1$}
                \STATE Observe a tuple $O_{k,m}^i = (s_{k,m}^i,a_{k,m}^i,r_{k,m}^i,s_{k,m + 1}^i)$ and estimate the advantage function
            \ENDFOR
            \STATE Update the local policy by (\ref{updatepolicy})
        \ENDFOR
        \STATE Server computes the global policy by (\ref{globalpolicy})
        \STATE \textbf{Output}: ${\theta _{k+1}}$
    \end{algorithmic}
\end{algorithm}
\\
\textbf{Algorithm summary.} In each outer loop $k \in \{ 1, \ldots , K\}$, the actor server first broadcasts the global policy $\pi_{\theta}$ to all agents. In inner loop communication round $t$, each critic $i \in \{ 1, \ldots ,N\}$ independently performs ${\upsilon_i}$ local iterations to approximate the value function of ${{\pi _\theta }}$ in their respective environments, as Algorithm 1 shows. Specifically, following the same policy ${{\pi _\theta }}$, agent $i$ observes the tuple $O_{k,t,\upsilon}^i =(s_{k,t,\upsilon}^i,r_{k,t,\upsilon}^i,s_{k,t,\upsilon+1}^i)$ at local iteration $\upsilon$ of round $t$ which is generated by its own MDP characterized by $\{ { S},{ A},{{ P}_i},{{ R}_i},{\gamma}\}$. Using observation $O_{k,t,\upsilon}^i$, agent $i$ computes the stochastic gradient and update its local model. At the end of each round, agents send the gradients directly to the critic server which then aggregates the gradients, updates the global critic model and starts round $t+1$ of federation. After $T$ rounds, the global critic parameter is sent to each actor. Then actors approximates the advantage function $A_{{\pi _\theta }}^i$ by the temporal difference error. The policy gradient can then be estimated as ${{\hat h}_t^i}(\omega_\theta ,\theta) = \delta _{\omega_\theta} ^i(s,a,s'){\psi _{\theta}}(s,a)$ as Algorithm 2 shows. Moreover, we employ Markovian mini-batch sampling to estimate the policy gradient, which helps reduce the variance in policy gradient estimation. Then the server aggregates local policy updates as a global policy and broadcasts it to all critics. 
\begin{algorithm}[!h]\label{FAC}
    \caption{Single-loop Federated Actor Critic (SFAC)}
    \label{alg:AOA}
    \renewcommand{\algorithmicrequire}{\textbf{Input:}}
    \renewcommand{\algorithmicensure}{\textbf{Output:}}
    \begin{algorithmic}[1]
        \STATE \textbf{Input}: number of  rounds $K$, stepsize $\alpha_k $, $\beta_k$, initial actor parameter ${{ \theta }_0}$
        \FOR{$k=1$ to $K$}
            \STATE $\omega_{k,0}=\omega_t$
            \FOR{$t  = 0$ to $T-1$}
                \STATE $\omega_{k,t+1}=\textbf{{FedC}}(\theta_t,\omega_{k,t},\beta_k)$
            \ENDFOR
            \STATE $\omega_{k+1}=\omega_{k,T}$
            \STATE $\theta_{k+1}=\textbf{{FedA}}(\theta_k,\omega_{k+1},\alpha_k)$    
        \ENDFOR
        \STATE \textbf{Output}: ${\theta _{\hat K}}$ with ${\hat K}$ chosen uniformly from $\{ 1, \ldots ,K\}$
    \end{algorithmic}
\end{algorithm}

\section{Convergence Analysis for SFAC}
In this section, we provide the convergence results of SFAC. To begin with, we make the following assumptions, which are commonly imposed in reinforcement learning and federated reinforcement learning \cite{fed_fault,fedhertd,fedtd,zeng2021decentralized, improving_liang}.

\begin{assumption} \textnormal{(Bounded Gradient Heterogeneity)}
For any set of weights satisfying convex combination, i.e., $\{ {p_i} \ge 0\} _{i = 1}^N$ and $\sum\nolimits_{i = 1}^N {{p_i}}  = 1$, there exist constants ${\chi ^2} \ge 1$, ${\kappa ^2} \ge 0$ such that $\sum\nolimits_i {{p_i}\left\| {{{g}_i}(\omega )} \right\|_2^2}  \le {\chi ^2}\left\| {\sum\nolimits_i {{p_i}} {{g}_i}(\omega )} \right\|_2^2 + {\kappa ^2}$. If agents are in identical environments, then ${\chi ^2} = 1$, ${\kappa ^2} = 0$.
\end{assumption}
Assumption 1 is commonly used in the federated learning literature to capture the dissimilarities of local objectives \cite{simfbo, nips20}.

\begin{assumption}
For each state-action pair $(s,a)$, any parameter $\theta$, $\theta ' \in {\mathbb{R}^d}$, the following inequalities always hold: $i){\left\| {{\psi _\theta }(s,a) - {\psi _{\theta '}}(s,a)} \right\|_2} \le {L_\psi }{\left\| {\theta  - \theta '} \right\|_2}$; $ii){\left\| {{\psi _\theta }(s,a)} \right\|_2} \le {C_\psi }$; $iii)\left| {{\pi _\theta }(a\left| s \right.) - {\pi _{\theta '}}(a\left| s \right.)} \right| \le {L_\pi }{\left\| {\theta  - \theta '} \right\|_2}$.
\end{assumption}

Assumption 2 holds for various commonly used policy classes\cite{konda1999actor,doya2000reinforcement}.  
\begin{assumption}\textnormal{(Ergodicity)}
For each $i\in [N]$, the Markov chain induced by policy ${{\pi _\theta }} $, corresponding to the state transition matrix ${ P}_i$, is aperiodic and irreducible. Then the geometric mixing property of the associated Markov chains is 
\begin{equation}
\sup {\left\| {{P_i}({x_t} \in  \cdot \left| {{x_0}} \right.) - {D _i}( \cdot )} \right\|_{TV}} \le {\eta _i}\rho _i^t
\end{equation}
where ${D_i ( \cdot )}$ is the stationary distribution of MDP $i$; ${\eta _i} > 0$ and ${\rho _i} \in [0,1]$ for all $i \in [N]$.
\end{assumption}
Assumption 3 is a common assumption \cite{PMLR2018,improving_liang,fedtd,fedsarsa} which holds for any uniformly ergodic Markov chain with general state spaces.

Now we are ready to present the main results of SFAC. First, we characterize the performance of the aggregated critics' model. The global objective function of our federated TD learning problem is the \textit{average MSPBE} of all agents for their respective environments, as the MSPBE quantifies the error of a value function model for an environment. 

\begin{proposition}\textnormal{For any policy $\pi_{\theta_k}$, $T$ represents the number of communication rounds for critics' federation. Consider FedC shown in Algorithm 2, assuming Assumptions 1 and 3 hold, we have:}
\begin{align*}
\mathbb{E}\left\| {{{ \omega  }_{k, T}} - \omega _{{\theta _k}}^*} \right\|_2^2 \le &\left( {1 - \frac{{\beta_k \overline \upsilon  \lambda }}{4}} \right)\left\| {{{ \omega  }_{k, T - 1}} - \omega _{{\theta _k}}^*} \right\|_2^2 \\
&+ {C_1}{\beta_k ^4} + {C_2}{\beta_k ^3} + {C_3}\frac{{{\beta_k ^2}}}{N} + {C_4}{\beta_k ^2}
\end{align*}
\textnormal{where $\lambda$, $C_1$, $C_2$, $C_3$ and $C_4$ are positive problem dependent constants and the detailed definitions of the constants are provided in the appendix. Note that when the heterogeneity level $\kappa^2=0$, $C_4$ will be zero.}

\end{proposition}

\begin{remark}
\textnormal{Proposition 1 provides a convergence error bound as $\left\| {{{\omega }_{k, T}} - \omega _{{\theta _k}}^*} \right\|_2^2$ of FedC, where ${{{\overline \omega  }_T}}$ is the global critic parameter. From Proposition 1, we can observe that the second and the third terms represent higher-order terms of step size $\beta_k$, which are negligible, compared to other terms. The fourth term captures the effect of noise where the variance gets scaled down by a factor of $N$ (linear speedup) due to collaboration among the agents. The fifth term describes the environmental heterogeneity.}
\end{remark}
Then we present the convergence results of SFAC as follows.
\begin{theorem}\textnormal{Consider SFAC, assuming Assumptions 1 to 3 hold, if we select ${\alpha _k} = O\left( {\sqrt {\frac{N}{K}} } \right)$, ${\beta _k} = O\left( {\sqrt {\frac{N}{K}} } \right)$, the output of Algorithm 3 satisfies:} 
\begin{align*}
&\frac{1}{K}\sum\limits_{k = 1}^K \mathbb{E}\left\| {\nabla J({\theta _k})} \right\|_2^2\\
&\le \frac{{2{V_1}}}{{{\alpha _k}K}} + \frac{{40{H^2}\left( {1 + (\eta  - 1)\rho } \right)}}{{M(1 - \rho )}} + \frac{{40{\kappa ^2}}}{{{c^2}}}+ {5{\xi _{critic}}} \\
&+2TC(\chi )C({\alpha _k})\left( {{C_1}\beta _k^2 + {C_2}{\beta _k} + \frac{{{C_3}}}{N} + {C_4}} \right)\frac{{\beta _k^2}}{{{\alpha _k}}}
\end{align*}
\textnormal{where we define the approximation error introduced by critics as ${\xi _{critic}} = \mathop {\max }\limits_{i,\theta} {E_{v_\theta^i}}{\left| {V_{{\pi _\theta}}^i(s) - V_{\omega _\theta^{i*}}^i(s)} \right|^2}$.}
\end{theorem}


\begin{remark}
\textnormal{Theorem 1 provides a bound on the convergence error of SFAC. The first term comes from the non-convex setting. The second term is the variance term which is controlled by the sample sizes of actors. The third term is determined by the heterogeneity level of environments. The last term shows how critics' error impact on the convergence bound. When the heterogeneity level equals to zero, the major term in the error bound ${\cal {O}}\left( {\frac{1}{{\sqrt {NK} }}} \right)$ shows a linear speedup.}
\end{remark}

\subsection{Proof Sketch of SFAC}
In this subsection, we present the key steps of the proof, and highlight the key technical differences in the convergence analysis of SFAC, compared with the previous works. 

In the convergence analysis, in order to study $\mathbb{E}\left\| {\nabla J\left( {{\theta_k}} \right)} \right\|_2^2$, we focus on analyzing the total variance term $\frac{1}{N}\sum\nolimits_{i = 1}^N {\hat h_k^i({\omega _{k + 1}},{\theta _k}) - \nabla J({\theta _k})} $ in (\ref{global_variance}) by five terms. The 1st term is an error introduced by the inaccurate estimations of the lower level. This term was directly bounded to zero under both the double-loop setting and the two-timescale setting due to their particular algorithm design, to enable a decoupled analysis. We make them as an extension of this paper later. The 2nd term "local variance" is a noise term introduced by Markovian sampling. This bias reduces to 0 under IID sampling after taking the expectation. The 3rd term is due to environment heterogeneity, which causes an non-vanishing term. The 4th term tracks the difference between the drifting critic targets, which is crucial to eliminate the error of critics. The 5th term is from the approximation error of the critics. 

\begin{align}\label{global_variance}
\quad&\left\| {\frac{1}{N}\sum\limits_{i = 1}^N {{{\hat h}_k^i}({{ \omega }_{k+1}},{\theta_k}) - \nabla J({\theta_k})} } \right\|_2^2 \nonumber\\
= &\left\| {\frac{1}{N}\sum\limits_{i = 1}^N \underbrace{{{{\hat h}_k^i}({{\omega }_{k + 1}},{\theta _k}) - {{\hat h}_k^i}({{\omega }^*_{k + 1}},{\theta _k})}}_{\text{error of lower level}}}  \right.\nonumber\\
& +\underbrace{{{\hat h}_k^i}({{\omega }^*_{k + 1}},{\theta _k}) - {h_k^i}(\omega _{k + 1}^*,{\theta _k})}_{\text{local variance}} + \underbrace{{h_k^i}(\omega _{k + 1}^*,{\theta _k}) - {h_k^i}(\omega _{k + 1}^{i*},{\theta _k})}_{\text{gradient heterogeneity}}\nonumber\\
&+\left. {\underbrace{{h_k^i}(\omega _{k + 1}^{i*},{\theta _k}) - {h_k^i}(\omega _k^{i*},{\theta _k})}_{\text{smoothness}} + \underbrace{{h_k^i}(\omega _k^{i*},{\theta _k}) - \nabla {J^i}({\theta _k})}_{\text{approximation error}}} \right\|_2^2\nonumber\\
\end{align}
\textbf{Error of lower level} Next, we present key steps and highlight the technical differences of Proposition 1, which provides bounds on the critic error. Similar to the convergence analysis of federated temporal difference learning \cite{fedtd, fedhertd,fedsarsa}, the contraction property of the Bellman equation is utilized to produce a descent direction for the critic error. The informal decomposition can be expressed as:
\\
$\mathbb{E}\left\| {{{\omega }_{t + 1}} - {\omega ^*}} \right\| \le  $ recursion + descent direction\\ \indent \quad \quad  + client drift + gradient variance + gradient norm.
\\
Note that we consider policy evaluation for policy $\pi_{\theta_k}$ and $\omega^*$ is the optimal value model for the mixture environment. In order to analyze 
$\mathbb{E}\left\| {{{\omega }_{k, t + 1}} - {\omega ^*}} \right\|$, we first bound an inner product term and it can be decomposed into three terms as shown in (\ref{tech_diff1}). As the objective of FedC algorithm is to minimize the average MSPBE, the term $B$ can be \textit{cancelled} (after the double summation before the inner product) due to the condition of the average of MSPBE. In contrast, in \cite{fedhertd,fedsarsa}, this term $B$ cannot be cancelled and becomes a \textit{non-vanishing bias} in the convergence error. Then, (\ref{tech_diff1}) contributes to the descent direction term and client drift term of the above decomposition. Furthermore, from Lemma 1, the descent direction provides the essential negative term so that the contraction property can be guaranteed.

\begin{align}\label{tech_diff1}
&\frac{1}{N}\sum\limits_i {\frac{{1}}{{\upsilon_i}}} \sum\limits_{\upsilon = 0}^{{\upsilon_i} - 1} {\mathbb{E}\left\langle {{{g}_i}(\omega _{k,t,v}^i),{{\omega_{k,t} }} - {\omega ^*}} \right\rangle } \nonumber\\
&= \frac{1}{N}\sum\limits_i {\frac{{1}}{{{\upsilon_i}}}} \sum\limits_{k = 0}^{{\upsilon_i} - 1} \mathbb{E}\left\langle {{{\omega_{k,t} }} - {\omega ^*}} \right.,\nonumber\\
&\left. {\underbrace {{{g}_i}(\omega _{k,t,v}^i) - {{g}_i}({\omega_{k,t} })}_{\text{client drift}} + \underbrace {{{g}_i}({{\omega_{k,t} }}) - g({\omega_{k,t} })}_B + \underbrace {g({\omega_{k,t} })}_{\text{decent direction}}} \right\rangle 
\end{align}
The gradient variance term and gradient norm term are commonly appear in FRL analyses. Specifically, the gradient variance term is controlled by the mixing property of the Markov chains \cite{markovbook}; the gradient norm term can be decomposed by the client drift and gradient variance terms, presenting \textit{linear speedup} effect as shown in Proposition 1.
\\
\textbf{Smoothness} Due to the Lipschitz continuity of $\omega^{i*}(\theta)$ as established in Lemma 3, the difference in the expected gradients at $\omega_{k+1}^{i*}$ and $\omega_k^{i*}$ can be related to the difference between $\theta_{k+1}$ and $\theta_k$. Since the two successive policies can be controlled by $\alpha_k$, the fourth term can also be controlled by $\alpha_k$. This is how the single-loop architecture works where the biased term in the critics can be eliminated in the actors.
\\
\textbf{Comparisons with double-loop AC} The proof addresses the policy drift issue in the critic update by considering the "lower-level error" in (\ref{global_variance}). When the policy changes, the critic's parameters are initialized from the parameters of the previous policy. However, the double-loop variant leverages this additional property to eliminate the term, enabling a decoupled analysis.

\section{Experiments}
We test the SFAC algorithm in the Lunar Lander environment and Cartpole environment provided by OpenAI Gym and the codes were running using T4 Tensor Core GPUs. We evaluate SFAC against \texttt{A3C} \cite{a3c}.

We use multilayer perceptrons for the critic model and the actor model. Specifically, the critic model uses a linear combination of the basis functions, composed of Relu functions of the weighted states, and the actor model uses a three-layer neural networks with Relu as the activation functions in the hidden layer and the softmax function in the output layer. The stepsizes are set to ${\alpha _k} = {10^{ - 4}} \times {0.99^{ - k}}$ and ${\beta _k} = {10^{ - 4}} \times {0.99^{ - k}}$. The discount factor is 0.99. We use multiple episodes to train the AC network. To measure the average performance, we collect the return for each episode and using the average of the latest 10 episodes for all agents as the average return during training. In each outer round, the number of local updating steps is set to $T= 10$. We sample $M= 20$ samples in each updating step of each inner round. Note that the term 'rounds' on the x-axis represents the number of iterations in the outer round.

\begin{figure}[!htbp]
    \centering
    \includegraphics[width=0.28\textwidth]{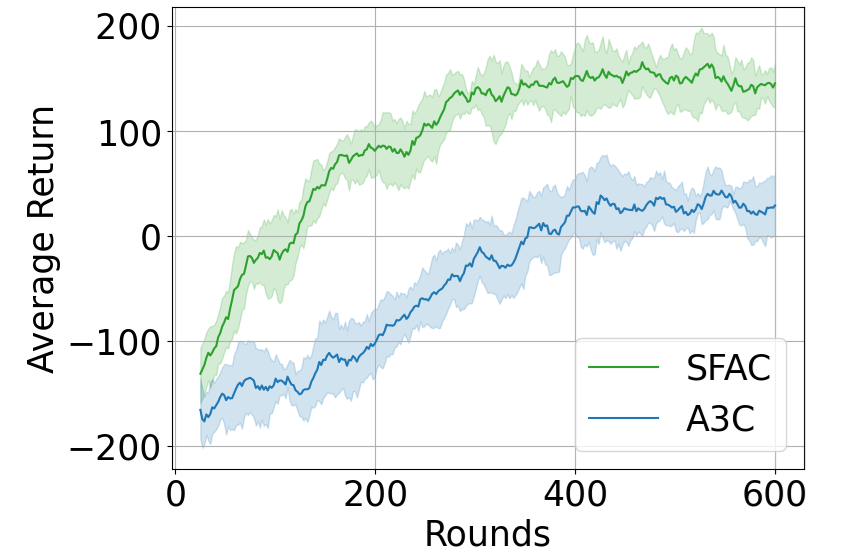}
    \vspace{-0.4cm}
    \caption{\centering SFAC Performance in Comparison to A3C}
    \label{C}
\end{figure}
\vspace{-1em}
\noindent
\textbf{Comparison with baseline.} We illustrate comparison results of SFAC with A3C. In terms of average return, SFAC manages to obtain higher objective values. Moreover, two-level federation accelerates the training process which matches our theoretical results.

\begin{figure}[!hb]
    \centering
    \begin{tikzpicture}
            \node[inner sep=0pt] (image) at (0,0) {\includegraphics[width=0.47\textwidth]{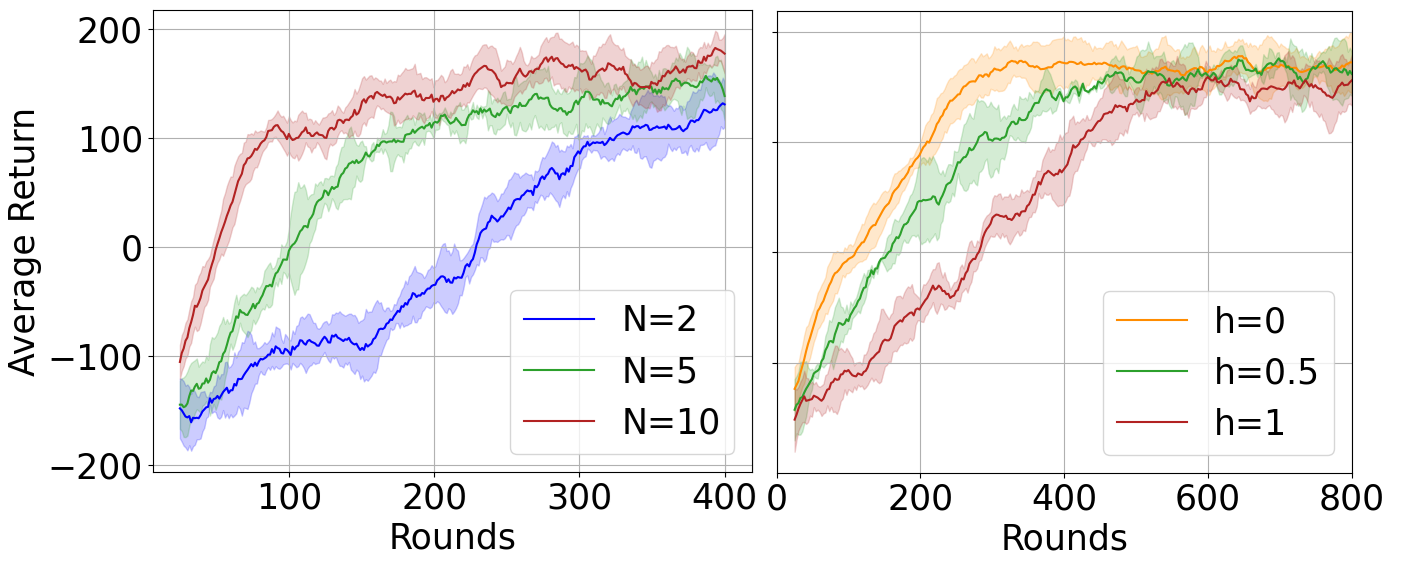}};
        \node[below=of image, yshift=1.0cm, xshift=-2.0cm] {\small (a) Number of agents};
        \node[below=of image, yshift=1.0cm, xshift=2.2cm] {\small (b) Environment heterogeneity};
    \end{tikzpicture} 
    \vspace{-0.8cm}
    \caption{Performance of SFAC}
    \label{P}
\end{figure}
\vspace{-1.2em}
\noindent
\textbf{Linear speedup.} To verify the advantages due to the federation, we conduct the experiments on the impact of the number of agents of SFAC. With a certain level of environment heterogeneity, increasing the number of participated agents accelerate the training. This corroborates our theoretical insights and verifies the practical performance benefit offered by the participation of more agents.
\\
\textbf{Environment heterogeneity.} To check the impact of environment heterogeneity on SFAC, we construct tasks of SFAC with various $h$, which controls how different the state transitions are. Fig. \ref{P} shows that, when we keep increasing $h$, the performance decreases. This result validates the theoretical results.

\section{Conclusion}
In this paper, we have studied SFAC with heterogeneous environments. We have designed a two-level collaboration algorithm where the critics perform federated TD learning for policy evaluation, while the actors perform federated policy update for policy improvement, to seek the optimal global policy across all environments. Furthermore, we have analyzed that SFAC can asymptotically produce a near stationary point with a linear speedup, which is the first result in existing works on FRL with actor-critic algorithms considering heterogeneous environments. 

\section{Acknowledgement}
This work was supported by U.S. NSF grants CNS-2145031, CNS-2206977.

\bibliography{./CameraReady/LaTeX/ac}

\begin{thebibliography}{36}
\providecommand{\natexlab}[1]{#1}

\bibitem[{Bhandari, Russo, and Singal(2018)}]{PMLR2018}
Bhandari, J.; Russo, D.; and Singal, R. 2018.
\newblock A finite time analysis of temporal difference learning with linear function approximation.
\newblock In \emph{Conference on learning theory (COLT)}.

\bibitem[{Chen, Sun, and Yin(2021)}]{alset}
Chen, T.; Sun, Y.; and Yin, W. 2021.
\newblock Closing the gap: Tighter analysis of alternating stochastic gradient methods for bilevel problems.
\newblock \emph{Advances in Neural Information Processing Systems}, 34: 25294--25307.

\bibitem[{Chen and Zhao(2024)}]{stac}
Chen, X.; and Zhao, L. 2024.
\newblock Finite-time analysis of single-timescale actor-critic.
\newblock \emph{Advances in Neural Information Processing Systems}, 36.

\bibitem[{Doan, Maguluri, and Romberg(2019)}]{TTD}
Doan, T.; Maguluri, S.; and Romberg, J. 2019.
\newblock Finite-time analysis of distributed TD (0) with linear function approximation on multi-agent reinforcement learning.
\newblock In \emph{International Conference on Machine Learning}, 1626--1635. PMLR.

\bibitem[{Doya(2000)}]{doya2000reinforcement}
Doya, K. 2000.
\newblock Reinforcement learning in continuous time and space.
\newblock \emph{Neural Computation}, 12(1): 219--245.

\bibitem[{Fan et~al.(2021)Fan, Ma, Dai, Jing, Tan, and Low}]{fed_fault}
Fan, X.; Ma, Y.; Dai, Z.; Jing, W.; Tan, C.; and Low, B. K.~H. 2021.
\newblock Fault-tolerant federated reinforcement learning with theoretical guarantee.
\newblock In \emph{Advances in Neural Information Processing Systems}, 1007--1021.

\bibitem[{Ghadimi and Wang(2018)}]{mengdi}
Ghadimi, S.; and Wang, M. 2018.
\newblock Approximation methods for bilevel programming.
\newblock \emph{arXiv preprint arXiv:1802.02246}.

\bibitem[{Hong et~al.(2023)Hong, Wai, Wang, and Yang}]{ttsa}
Hong, M.; Wai, H.-T.; Wang, Z.; and Yang, Z. 2023.
\newblock A two-timescale stochastic algorithm framework for bilevel optimization: Complexity analysis and application to actor-critic.
\newblock \emph{SIAM Journal on Optimization}, 33(1): 147--180.

\bibitem[{Huang, Zhang, and Ji(2023)}]{fedmbo}
Huang, M.; Zhang, D.; and Ji, K. 2023.
\newblock Achieving linear speedup in non-iid federated bilevel learning.
\newblock In \emph{International Conference on Machine Learning}, 14039--14059. PMLR.

\bibitem[{Jin et~al.(2022)Jin, Peng, Yang, Wang, and Zhang}]{aistat22}
Jin, H.; Peng, Y.; Yang, W.; Wang, S.; and Zhang, Z. 2022.
\newblock Federated reinforcement learning with environment heterogeneity.
\newblock In \emph{International Conference on Artificial Intelligence and Statistics}, 18--37. PMLR.

\bibitem[{Khodadadian et~al.(2022)Khodadadian, Sharma, Joshi, and Maguluri}]{fedtd}
Khodadadian, S.; Sharma, P.; Joshi, G.; and Maguluri, S.~T. 2022.
\newblock Federated reinforcement learning: Linear speedup under markovian sampling.
\newblock In \emph{International Conference on Machine Learning}, 10997--11057. PMLR.

\bibitem[{Kiran et~al.(2021)Kiran, Sobh, Talpaert, Mannion, Al~Sallab, Yogamani, and P{\'e}rez}]{auto_driving_2}
Kiran, B.~R.; Sobh, I.; Talpaert, V.; Mannion, P.; Al~Sallab, A.~A.; Yogamani, S.; and P{\'e}rez, P. 2021.
\newblock Deep reinforcement learning for autonomous driving: A survey.
\newblock \emph{IEEE Transactions on Intelligent Transportation Systems}, 23(6): 4909--4926.

\bibitem[{Konda and Borkar(1999)}]{konda1999actor}
Konda, V.~R.; and Borkar, V.~S. 1999.
\newblock Actor-critic--type learning algorithms for Markov decision processes.
\newblock \emph{SIAM Journal on Control and Optimization}, 38(1): 94--123.

\bibitem[{Kumar, Koppel, and Ribeiro(2023)}]{kumar2023sample}
Kumar, H.; Koppel, A.; and Ribeiro, A. 2023.
\newblock On the sample complexity of actor-critic method for reinforcement learning with function approximation.
\newblock \emph{Machine Learning}, 112(7): 2433--2467.

\bibitem[{Levin and Peres(2017)}]{markovbook}
Levin, D.~A.; and Peres, Y. 2017.
\newblock \emph{Markov chains and mixing times}, volume 107.
\newblock American Mathematical Soc.

\bibitem[{Qiu et~al.(2021)Qiu, Yang, Ye, and Wang}]{qiu}
Qiu, S.; Yang, Z.; Ye, J.; and Wang, Z. 2021.
\newblock On finite-time convergence of actor-critic algorithm.
\newblock \emph{IEEE Journal on Selected Areas in Information Theory}, 2(2): 652--664.

\bibitem[{Shalev-Shwartz, Shammah, and Shashua(2016)}]{auto_driving_1}
Shalev-Shwartz, S.; Shammah, S.; and Shashua, A. 2016.
\newblock Safe, multi-agent, reinforcement learning for autonomous driving.
\newblock \emph{arXiv preprint arXiv:1610.03295}.

\bibitem[{Shen et~al.(2023)Shen, Zhang, Hong, and Chen}]{a3c}
Shen, H.; Zhang, K.; Hong, M.; and Chen, T. 2023.
\newblock Towards understanding asynchronous advantage actor-critic: Convergence and linear speedup.
\newblock \emph{IEEE Transactions on Signal Processing}, 71: 2579--2594.

\bibitem[{Srikant and Ying(2019)}]{srikant19colt}
Srikant, R.; and Ying, L. 2019.
\newblock Finite-time error bounds for linear stochastic approximation andtd learning.
\newblock In \emph{Conference on Learning Theory (COLT)}.

\bibitem[{Sun et~al.(2020)Sun, Wang, Giannakis, Yang, and Yang}]{sun2020finite}
Sun, J.; Wang, G.; Giannakis, G.~B.; Yang, Q.; and Yang, Z. 2020.
\newblock Finite-time analysis of decentralized temporal-difference learning with linear function approximation.
\newblock In \emph{International Conference on Artificial Intelligence and Statistics (AISTATS)}.

\bibitem[{Sutton et~al.(1999)Sutton, McAllester, Singh, and Mansour}]{sutton99}
Sutton, R.~S.; McAllester, D.~A.; Singh, S.~P.; and Mansour, Y. 1999.
\newblock Policy gradient methods for reinforcement learning with function approximation.
\newblock In \emph{Advances in Neural Information Processing Systems}, 1057--1063.

\bibitem[{Tarzanagh et~al.(2022)Tarzanagh, Li, Thrampoulidis, and Oymak}]{fednest}
Tarzanagh, D.~A.; Li, M.; Thrampoulidis, C.; and Oymak, S. 2022.
\newblock Fednest: Federated bilevel, minimax, and compositional optimization.
\newblock In \emph{International Conference on Machine Learning}, 21146--21179. PMLR.

\bibitem[{Wang et~al.(2023)Wang, Mitra, Hassani, Pappas, and Anderson}]{fedhertd}
Wang, H.; Mitra, A.; Hassani, H.; Pappas, G.~J.; and Anderson, J. 2023.
\newblock Federated temporal difference learning with linear function approximation under environmental heterogeneity.
\newblock \emph{arXiv preprint arXiv:2302.02212}.

\bibitem[{Wang et~al.(2020)Wang, Liu, Liang, Joshi, and Poor}]{nips20}
Wang, J.; Liu, Q.; Liang, H.; Joshi, G.; and Poor, H.~V. 2020.
\newblock Tackling the Objective Inconsistency Problem in Heterogeneous Federated Optimization.
\newblock In \emph{Advances in Neural Information Processing Systems (NIPS)}.

\bibitem[{Wang et~al.(2019)Wang, Cai, Yang, and Wang}]{PGgloptimal}
Wang, L.; Cai, Q.; Yang, Z.; and Wang, Z. 2019.
\newblock Neural policy gradient methods: Global optimality and rates of convergence.
\newblock \emph{arXiv preprint arXiv:1909.01150}.

\bibitem[{Williams(1992)}]{williams}
Williams, R.~J. 1992.
\newblock Simple statistical gradient-following algorithms for connectionist reinforcement learning.
\newblock \emph{Machine learning}, 8: 229--256.

\bibitem[{Xie and Song(2023)}]{fedkl}
Xie, Z.; and Song, S. 2023.
\newblock FedKL: Tackling data heterogeneity in federated reinforcement learning by penalizing KL divergence.
\newblock \emph{IEEE Journal on Selected Areas in Communications}, 41(4): 1227--1242.

\bibitem[{Xu, Wang, and Liang(2020)}]{improving_liang}
Xu, T.; Wang, Z.; and Liang, Y. 2020.
\newblock Improving sample complexity bounds for (natural) actor-critic algorithms.
\newblock \emph{Advances in Neural Information Processing Systems (NeurIPS)}.

\bibitem[{Yang, Xiao, and Ji(2024)}]{simfbo}
Yang, Y.; Xiao, P.; and Ji, K. 2024.
\newblock Simfbo: Towards simple, flexible and communication-efficient federated bilevel learning.
\newblock \emph{Advances in Neural Information Processing Systems}, 36.

\bibitem[{Ye, Li, and Juang(2019)}]{resource1}
Ye, H.; Li, G.~Y.; and Juang, B.-H.~F. 2019.
\newblock Deep reinforcement learning based resource allocation for V2V communications.
\newblock \emph{IEEE Transactions on Vehicular Technology}, 68(4): 3163--3173.

\bibitem[{Yu et~al.(2020)Yu, Chen, Zhou, Gong, and Wu}]{resource2}
Yu, S.; Chen, X.; Zhou, Z.; Gong, X.; and Wu, D. 2020.
\newblock When deep reinforcement learning meets federated learning: Intelligent multitimescale resource management for multiaccess edge computing in 5G ultradense network.
\newblock \emph{IEEE Internet of Things Journal}, 8(4): 2238--2251.

\bibitem[{Zeng et~al.(2021)Zeng, Anwar, Doan, Raychowdhury, and Romberg}]{zeng2021decentralized}
Zeng, S.; Anwar, M.~A.; Doan, T.~T.; Raychowdhury, A.; and Romberg, J. 2021.
\newblock A decentralized policy gradient approach to multi-task reinforcement learning.
\newblock In \emph{Uncertainty in Artificial Intelligence (UAI)}, 1002--1012. PMLR.

\bibitem[{Zhang et~al.(2024)Zhang, Wang, Mitra, and Anderson}]{fedsarsa}
Zhang, C.; Wang, H.; Mitra, A.; and Anderson, J. 2024.
\newblock Federated temporal difference learning with linear function approximation under environmental heterogeneity.
\newblock In \emph{International Conference on Learning Representations (ICLR)}.

\bibitem[{Zhang, Yang, and Ba{\c{s}}ar(2021)}]{zhang2021decentralized}
Zhang, K.; Yang, Z.; and Ba{\c{s}}ar, T. 2021.
\newblock Decentralized multi-agent reinforcement learning with networked agents: Recent advances.
\newblock \emph{Frontiers of Information Technology \& Electronic Engineering}, 22(6): 802--814.

\bibitem[{Zhang et~al.(2018)Zhang, Yang, Liu, Zhang, and Basar}]{zhang2018fully}
Zhang, K.; Yang, Z.; Liu, H.; Zhang, T.; and Basar, T. 2018.
\newblock Fully decentralized multi-agent reinforcement learning with networked agents.
\newblock In \emph{International Conference on Machine Learning (ICML)}, 5872--5881. PMLR.

\bibitem[{Zhu and Gong(2023)}]{ciss}
Zhu, Y.; and Gong, X. 2023.
\newblock Distributed policy gradient with heterogeneous computations for federated reinforcement learning.
\newblock In \emph{2023 57th Annual Conference on Information Sciences and Systems (CISS)}, 1--6. IEEE.

\end{thebibliography}

\newpage
\onecolumn
\section{Supplementary Materials}
We will provide the full version proof of Proposition 1 and Theorem 1.

\subsection{Key Lemmas}

\begin{lemma}\label{lemma2}
There exists a Markov chain with expected gradient calculated by temporal difference learning is $g(\omega )$. Then it holds that
\begin{align}
g(\omega ) &= \frac{1}{N}\sum\limits_i {{{g}_i}(\omega ) }\nonumber\\
&=  \frac{1}{N}\sum\limits_i {\left( {{\Phi ^{\rm{T}}}{D_i}(\gamma {P_i}\Phi  - \Phi )\omega  + {\Phi ^{\rm{T}}}{D_i}{R_i}} \right)}\nonumber \\
& = {\Phi ^{\rm{T}}}{D^*}{R^*} + {\Phi ^{\rm{T}}}{D^*}(\gamma {P^*}\Phi- \Phi )\omega\nonumber
\end{align}
\begin{proof} \label{lemma2}
The sufficient condition is 
\begin{align}  \label{19}
\left\{ {\begin{array}{*{20}{c}}
\frac{1}{N}\sum\limits_i {{D_i}{R_i} = } {D^*}{R^*}\\
\frac{1}{N}\sum\limits_i {{D_i}(\gamma {P_i} - I) = {D^*}(\gamma {P^*} - I)}
\end{array}} \right.
\end{align}
where $I$ is the identity matrix. Then we can find one solution of (\ref{19}) as 
\begin{align*}
\left\{ {\begin{array}{*{20}{c}}
{{R^*} = {{\left( {\frac{1}{N}\sum\limits_i {{D_i}} } \right)}^{ - 1}}\left( {\frac{1}{N}\sum\limits_i {{D_i}{R_i}} } \right)}\\
{{P^*} = {{\left( {\frac{1}{N}\sum\limits_i {{D_i}} } \right)}^{ - 1}}\left( {\frac{1}{N}\sum\limits_i {{D_i}{P_i}} } \right)}
\end{array}} \right.
\end{align*}
where we assume ${\frac{1}{N}\sum\limits_i {{D_i}} }$ is invertible. Since $D_i$ is a diagonal matrix, it can be easily proved that ${P^*}$ is a transition matrix which is irreducible. 
\end{proof}
\end{lemma}

Lemma 1 lays the groundwork for Assumption \ref{liang} to hold, which is important for facilitating the convergence analysis.

\begin{assumption}\textnormal{(Assumption 3 in \cite{improving_liang})} \label{liang}
Assume that $\left\| {\phi (s)} \right\|_2^2 \le 1$ for all $s$ and the columns of the feature matrix $\Phi $ are linearly independent. Then, there exists a positive number $\lambda$ such that
\begin{align*}
\left\langle { g(\omega ),\omega  - {\omega ^*}} \right\rangle  \le  - \frac{\lambda }{2}\left\| {\omega  - {\omega ^*}} \right\|_2^2. 
\end{align*}
\end{assumption}
\noindent
\textbf{Proof of Proposition 1}
Inspired by \cite{sun2020finite,fedtd,fedhertd}, we first decompose the stochastic gradient as
\begin{equation*}
{\hat g_i}(\omega _{t,\upsilon}^i) = {\hat A_i}(O_{t,\upsilon}^i)(\omega _{t,\upsilon}^i - \omega _i^*) + \underbrace {{\hat A_i}(O_{t,\upsilon}^i)\omega _i^* + {\hat b_i}\left( {O_{t,\upsilon}^i} \right)}_{{\hat Z_i}(O_{t,\upsilon}^i)} 
\end{equation*}
From the analysis in IID setting, we know that the expected gradient can be written as
\begin{equation}
{\hat{g}_i}(\omega _{t,\upsilon}^i) = {\hat A_i}(O_{t,\upsilon}^i)(\omega _{t,\upsilon}^i - \omega _i^*)    
\end{equation}
where ${{ A}_i} = {\Phi ^{\rm{T}}}{D_i}(\gamma {P_i}\Phi  - \Phi )$. At the same time, ${{\mathbb{E}}_{O_{t,\upsilon}^i\sim {\pi _i}}}\left[ {{\hat Z_i}(O_{t,\upsilon}^i)} \right] = 0$ holds.

In the following, We first give the uniform bound of these variables.
\begin{lemma}(\textbf{Bounded Variables})
\label{lemma5}\\
Given $\left\| {\phi (s)} \right\|_2^2 \le 1$, we can bound the following variables as ${\left\| {{\hat A_i}(O_{t,\upsilon}^i)} \right\|_2} \le 1 + \gamma $, ${\left\| {{{ A}_i}(O_{t,\upsilon}^i)} \right\|_2} \le 1 + \gamma $ and ${\left\| {{\hat Z_i}(O_{t,\upsilon}^i)} \right\|_2} \le ( 1 + \gamma)H + R$ hold for all $i \in [N]$, where $H$ denotes the radius of the set which includes all local optimal solutions and global solution and $R$ denotes the upper bound of the reward.
\begin{proof}
Since $\left\| {\phi (s)} \right\|_2^2 \le 1$, we can conclude that
\begin{equation*}
{\left\| {{\hat A_i}(O_{t,\upsilon}^i)} \right\|_2} = {\left\| {\phi (s_{t,\upsilon}^i)(\gamma {\phi ^{\rm T}}(s_{t,\upsilon}^i) - {\phi ^{\rm T}}(s_{t,\upsilon+1}^i))} \right\|_2} \le {\left\| {\phi (s_{t,\upsilon}^i)} \right\|_2}{\left\| {\gamma {\phi ^{\rm T}}(s_{t,\upsilon}^i) - {\phi ^{\rm T}}(s_{t,\upsilon+1}^i)} \right\|_2} \le c    
\end{equation*}  
where $c: = 1 + \gamma $. Then using the triangle inequality, we have:
\begin{equation}
 {\left\| {{\hat Z_i}(O_{t,\upsilon}^i)} \right\|_2} = {\left\| {{\hat A_i}(O_{t,\upsilon}^i)\omega _i^* + {\hat b_i}(O_{t,\upsilon}^i)} \right\|_2} \le {\left\| {{\hat A_i}(O_{t,\upsilon}^i)\omega _i^*} \right\|_2} + {\left\| {{\hat b_i}(O_{t,\upsilon}^i)} \right\|_2} \le cH + R   
\end{equation}
\end{proof}
\end{lemma}

Since there are strong dependencies between the noisy observations in the Markovian chain model \cite{PMLR2018} , we need to compute the bias between the expectation of stochastic $A_i$ and ${{{ A}_i}}$ which is zero in IID sampling, and the bias of the expectation of $Z_i$ which is also zero in IID sampling.
\begin{lemma}(\textbf{Bias of Variables' Expectation}) \label{lemma6}
\begin{equation*}
{\left\| {{{ A}_i} -  \mathbb{E}\left[ {{A_i}(O_i^{{t_2},{k_2}})\left| {F_{{k_1}}^{{t_1}}} \right.} \right]} \right\|_2} \le 2c{\eta _i}\rho _i^{\sum\limits_{t = {t_1}}^{{t_2} - 1} {{\upsilon_i}}  + {k_2} - {k_1}} \le c'\rho _i^{\sum\limits_{t = {t_1}}^{{t_2} - 1} {{\upsilon_i}}  + {k_2} - {k_1}}
\end{equation*}
where $c' = 2c\mathop {\max }\limits_i \left\{ {{\eta _i}} \right\}$

\end{lemma}

Similarly, for the bias of the expectation of $Z_i$ using Markovian sampling, we have

\begin{equation*}
{\left\| { \mathbb{E}\left[ {{Z_i}(O_i^{{t_2},{k_2}})\left| {F_{{k_1}}^{{t_1}}} \right.} \right]} \right\|_2} \le 2q{\eta _i}\rho _i^{\sum\limits_{t = {t_1}}^{{t_2} - 1} {{\upsilon_i}}  + {k_2} - {k_1}} \le q'\rho _i^{\sum\limits_{t = {t_1}}^{{t_2} - 1} {{\upsilon_i}}  + {k_2} - {k_1}}
\end{equation*}
where $q: = cH + R$; $q' = 2q\mathop {\max }\limits_i \left\{ {{\eta _i}} \right\}$.
The proof can be found in \cite{sun2020finite,fedhertd}.

\begin{lemma}
${J_i}(\theta )$ is $L_i$-smooth and the aggregated value function $J(\theta )$ is $L$-smooth with $L = \sum\nolimits_{i = 1}^N {{L_i}} $
\end{lemma}

\begin{lemma}\textnormal{(Proposition 7 in \cite{alset})}\label{omega}
Suppose Assumption 2-4 hold. For any $\theta_1, \theta_2 \in \mathbb{R}^d$, we have 
\begin{align}
{\left\| {{\omega ^*}({\theta _1}) - {\omega ^*}({\theta _2})} \right\|_2} \le {L_\omega }{\left\| {{\theta _1} - {\theta _2}} \right\|_2}
\end{align}
where $L_\omega$ is a positive constant.
\end{lemma}

\begin{lemma}\textnormal{(Proposition 8 in \cite{alset})} \label{nomega}
Suppose Assumption 2-4 hold. For any $\theta_1, \theta_2 \in \mathbb{R}^d$, we have 
\begin{align}
{\left\| {{\nabla _\theta }{\omega ^*}({\theta _1}) - {\nabla _\theta }{\omega ^*}({\theta _2})} \right\|_2} \le {L_{\omega \theta }}{\left\| {{\theta _1} - {\theta _2}} \right\|_2}
\end{align}
where $L_{\omega \theta }$ is a positive constant.
\end{lemma}

\begin{lemma}\textnormal{(Adapted from \cite{improving_liang}) }
\begin{equation*}
\mathbb{E}\left\| {{{\hat h}^i}(\omega _{{\theta_k}}^*,{\theta_k}) - {h^i}(\omega _{{\theta_k}}^*,{\theta_k})} \right\|_2^2 \le \frac{{8{H^2}\left( {1 + ({\eta _i} - 1){\rho _i}} \right)}}{{M(1 - {\rho _i})}}
\end{equation*}
\end{lemma}

\subsection{Variance reduction}
In this section, we give the bound of the gradient variance. Since the gradient can be decomposed into two terms, we discuss them separately. 

\begin{align}
&{ \mathbb{E}_{t - \tau }}\left\| {\frac{1}{N}\sum\limits_i {\frac{{1}}{{{\upsilon_i}}}\sum\limits_{k = 0}^{{\upsilon_i-1}} {{\hat Z_i}(O_{t,\upsilon}^i)} } } \right\|_2^2\nonumber\\
 = &{ \mathbb{E}_{t - \tau }}{\left( {\frac{1}{N}\sum\limits_i {\frac{{1}}{{{\upsilon_i}}}\sum\limits_{k = 0}^{{\upsilon_i-1}} {{\hat Z_i}(O_{t,\upsilon}^i)} } } \right)^{\rm T}}\left( {\frac{1}{N}\sum\limits_i {\frac{{1}}{{{\upsilon_i}}}\sum\limits_{k = 0}^{{\upsilon_i-1}} {{\hat Z_i}(O_{t,\upsilon}^i)} } } \right)\nonumber\\
 = &\frac{1}{{{N^2}}}{\mathbb{E}_{t - \tau }}\left[ {\sum\limits_i {\frac{1}{{{{\upsilon _i^2}}}}\sum\limits_{k = 0}^{{\upsilon _i} - 1} {\hat Z_i^{\rm{T}}(O_{t,\upsilon }^i)} \hat Z_i^{\rm{T}}(O_{t,\upsilon }^i) + 2\sum\limits_i {\frac{1}{{{{\upsilon _i^2}}}}\sum\limits_{k < l} {\hat Z_i^{\rm{T}}(O_{t,\upsilon }^i){\hat Z_i}(O_{t,l}^i)} } } } \right.\nonumber\\
&\left. { + 2\sum\limits_{i < j} {\frac{1}{{{\upsilon _i}{\upsilon _j}}}\sum\limits_{k \in [{\upsilon _i}],l \in [{\upsilon _i}]} {\hat Z_i^{\rm{T}}(O_{t,\upsilon }^i){\hat Z_j}(O_{t,l}^j)} } } \right] \label{vr}
\end{align}

For the second term of (\ref{vr}),
\begin{align*}
&{ \mathbb{E}_{t - \tau }}\left[ {2\sum\limits_i {\frac{1}{{{{\upsilon _i^2}}}}\sum\limits_{k < l} {\hat Z_i^{\rm{T}}(O_{t,\upsilon}^i){\hat Z_i}(O_{t,l}^i)} } } \right]\\
= &{ \mathbb{E}_{t - \tau }}\left[ {2\sum\limits_i {\frac{1}{{{{\upsilon _i^2}}}}\sum\limits_{k < l} {\hat Z_i^{\rm{T}}(O_{t,\upsilon}^i) \mathbb{E}\left[ {{\hat Z_i}(O_{t,l}^i)\left| {F_t^k} \right.} \right]} } } \right]\\
\le &{ \mathbb{E}_{t - \tau }}\left[ {2qq'\sum\limits_i {\frac{1}{{{{\upsilon _i^2}}}}\sum\limits_{k < l} {{\rho ^{l - k}}} } } \right]
\end{align*}
where the last inequality is derived from Lemma \ref{lemma5} and Lemma \ref{lemma6}. Then it follows:
\begin{align}
{ \mathbb{E}_{t - \tau }}\left[ {2\sum\limits_i {\frac{1}{{{{{\upsilon _i^2}}}}}\sum\limits_{k < l} {\hat Z_i^{\rm{T}}(O_{t,\upsilon}^i){\hat Z_i}(O_{t,l}^i)} } } \right]\le &{ \mathbb{E}_{t - \tau }}\left[ {2qq'\sum\limits_i {\frac{1}{{{{{\upsilon _i^2}}}}}\sum\limits_{\upsilon = 0}^{{\upsilon_i} - 1} {\sum\limits_{l = k + 1}^{{\upsilon_i} - 1} {\rho ^{l - k}} } } } \right]\nonumber\\
\le &{ \mathbb{E}_{t - \tau }}\left[ {\frac{{2qq'\rho }}{{1 - \rho }}\sum\limits_i {\frac{{{\upsilon_i} - 1}}{{{{{\upsilon _i^2}}}}}} } \right] \label{41}\\
\le &{ \mathbb{E}_{t - \tau }}\left[ {\frac{{2qq'\rho ({\upsilon_{\max}}-1)}}{{1 - \rho }}\sum\limits_i {\frac{1}{{{{{\upsilon _i^2}}}}}} } \right]\nonumber
\end{align}
where $\rho  = \mathop {\max }\limits_i \left\{ {{\rho _i}} \right\}$, ${\upsilon_{\max}} = \mathop {\max }\limits_i \{ {\upsilon_i}\} $ and (\ref{41}) is derived from the fact that $\sum\limits_{a = 1}^\infty  {{\rho ^a} = \frac{\rho }{{1 - \rho }}} $ when $0 < \rho  < 1$. Now we study the third term of (\ref{vr}) where observation of different agents are mutually independent.

For the third term of (\ref{vr}),
\begin{align*}
&{ \mathbb{E}_{t - \tau }}\left[ {2\sum\limits_{i < j} {\frac{1}{{{\upsilon_i}{\upsilon_j}}}\sum\limits_{k \in [{\upsilon_i}],l \in [{\upsilon_j}]} {\hat Z_i^{\rm{T}}(O_{t,\upsilon}^i){\hat Z_j}(O_j^{t,l})} } } \right]\\
 \le& 2\sum\limits_{i < j} {\frac{1}{{{\upsilon_i}{\upsilon_j}}}\sum\limits_{k \in [{\upsilon_i}],l \in [{\upsilon_j}]} {{ \mathbb{E}_{t - \tau }}{\left\| {{\hat Z_i^{\rm{T}}(O_{t,j}^i)}} \right\|_2}
{ \mathbb{E}_{t - \tau }}{\left\| {{{\hat Z_k}(O_{t,l}^k))}} \right\|_2}} } \\
 =& 2{[q']^2}\sum\limits_{i < j} {\frac{1}{{{\upsilon_i}{\upsilon_j}}}\sum\limits_{k \in [{\upsilon_i}],l \in [{\upsilon_j}]} {{\rho ^{\tau(\upsilon_i+\upsilon_j)+k + l}}} } 
\end{align*}
where the last inequality is derived from Lemma \ref{lemma6}.
Then the third term of (\ref{vr}) can be further bounded as
\begin{align*}
&{\mathbb{E}_{t - \tau }}\left[ {2\sum\limits_{i < j} {\frac{1}{{{\upsilon_i}{\upsilon_j}}}\sum\limits_{k \in [{\upsilon_i}],l \in [{\upsilon_j}]} {\hat Z_i^{\rm{T}}(O_{t,\upsilon}^i){\hat Z_j}(O_{t,l}^j)} } } \right]\\
 = &2{[q']^2}\sum\limits_{i < j} {\frac{1}{{{\upsilon_i}{\upsilon_j}}}{\rho ^{\tau ({\upsilon_i} + {\upsilon_j})}}} \sum\limits_{k \in [{\upsilon_i}],l \in [{\upsilon_j}]} {{\rho ^{k + l}}} \\
 \le& \frac{{2{{[q']}^2}\rho ({\upsilon_{\max}} - 1)}}{{1 - \rho }}\sum\limits_{i < j} {\frac{1}{{{\upsilon_i}{\upsilon_j}}}{\rho ^{\tau(\upsilon_i+\upsilon_j)}}} 
\end{align*}

Then (\ref{vr}) can be represented as 
\begin{align} 
{\mathbb{E}_{t - \tau }}\left\| {\frac{1}{N}\sum\limits_i {\frac{{1}}{{{\upsilon_i}}}\sum\limits_{\upsilon = 0}^{{\upsilon_i} - 1} {{\hat Z_i}(O_{t,\upsilon}^i)} } } \right\|_2^2 &\le \frac{1}{{{N^2}}}\sum\limits_i {\frac{{{{[q']}^2}}}{{{\upsilon_i}}}}  + \frac{{2qq'\rho ({\upsilon_{\max}} - 1)}}{{{N^2}}{(1 - \rho) }}\sum\limits_i {\frac{1}{{{{{\upsilon _i^2}}}}}} \nonumber\\
& + \frac{{2{{[q']}^2}\rho ({\upsilon_{\max}} - 1)}}{{{N^2}}{(1 - \rho) }}\sum\limits_{i < j} {\frac{1}{{{\upsilon_i}{\upsilon_j}}}{\rho ^{\tau(\upsilon_i+\upsilon_j)}}} \label{vro}
\end{align}

Similarly, we have
\begin{align}
{\mathbb{E}_{t - \tau }}\left\| {\frac{1}{N}\sum\limits_i {\frac{{1}}{{{\upsilon_i}}}\sum\limits_{\upsilon = 0}^{{\upsilon_i} - 1} {{\hat A_i}(O_{t,\upsilon}^i) - {{ A}_i}} } } \right\|_2^2 &\le \frac{1}{{{N^2}}}\sum\limits_i {\frac{{1{{[c']}^2}}}{{{\upsilon_i}}}}  + \frac{{2cc'\rho ({\upsilon_{\max}} - 1)}}{{{{N^2}}{(1 - \rho) }}}\sum\limits_i {\frac{1}{{{{{\upsilon _i^2}}}}}} \nonumber\\
& + \frac{{2{{[c']}^2}\rho ({\upsilon_{\max}} - 1)}}{{{N^2}}{(1 - \rho) }}\sum\limits_{i < j} {\frac{1}{{{\upsilon_i}{\upsilon_j}}}{\rho ^{\tau(\upsilon_i+\upsilon_j)}}} 
\end{align}

\subsection{Per Round Progress in Markovian Sampling}

In this section, we will analyze the convergence performance under Markovian sampling. For any $t>\tau$, we have

\begin{align} \label{Per_mar}
&{\mathbb{E}_{t - \tau }}\left\| {{{\bar \omega }_{t + 1}} - {\omega ^*}} \right\|_2^2 \nonumber\\
= &{\mathbb{E}_{t - \tau }}\left\| {{{{\bar \omega }_t} + \beta_t {\bar {\upsilon}} \cdot \frac{1}{N}\sum\limits_i {d_{t}^i}  - {\omega ^*}}} \right\|_2^2\nonumber\\
 = &{\mathbb{E}_{t - \tau }}\left\| {{{\bar \omega }_t} - {\omega ^*}} \right\|_2^2 + {2\beta_t {\bar {\upsilon}}}\underbrace { {\mathbb{E}_{t - \tau }}\left\langle {\frac{1}{N}\sum\limits_i {f_{t}^i} ,{{\bar \omega }_t} - {\omega ^*}} \right\rangle }_{{W_1}} + {\beta_t ^2}{{\bar {\upsilon}}^2}\underbrace {{\mathbb{E}_{t - \tau }}\left\| {\frac{1}{N}\sum\limits_i {d_{t}^i} } \right\|_2^2}_{{W_2}}\nonumber\\
&+ 2\beta_t {\bar {\upsilon}}\underbrace {{\mathbb{E}_{t - \tau }}\left\langle {\frac{1}{N}\sum\limits_i {(d_{t}^i - f_{t}^i)} ,{{\bar \omega }_t} - {\omega ^*}} \right\rangle }_{{W_3}}
\end{align}
where $W_3$ is a biased term due to Markovian sampling.

For the second term $W_1$ on the right hand side in (\ref{Per_mar}), we can obtain:

\begin{align}
&{W_1} =  {\mathbb{E}_{t - \tau }}\left\langle {\frac{1}{N}\sum\limits_i {f_{t}^i} ,{{\bar \omega }_t} - {\omega ^*}} \right\rangle  =  {\mathbb{E}_{t - \tau }}\left\langle {\frac{1}{N}\sum\limits_i {\frac{{1}}{{{\upsilon_i}}}\sum\limits_{\upsilon = 0}^{{\upsilon_i} - 1} {{{ g}_i}(\omega _{t,\upsilon}^i),{{\bar \omega }_t} - {\omega ^*}} } } \right\rangle \nonumber\\
 = & {\mathbb{E}_{t - \tau }}\left\langle {\frac{1}{N}\sum\limits_i {\frac{{1}}{{{\upsilon_i}}}\sum\limits_{\upsilon = 0}^{{\upsilon_i} - 1} {({{ g}_i}(\omega _{t,\upsilon}^i) - {{ g}_i}({{\bar \omega }_t}) + {{ g}_i}({{\bar \omega }_t}) - g({{\bar \omega }_t}) + g({{\bar \omega }_t})),{{\bar \omega }_t} - {\omega ^*}} } } \right\rangle \nonumber\\
 = & {\mathbb{E}_{t - \tau }}\left\langle {\frac{1}{N}\sum\limits_i {\frac{{1}}{{{\upsilon_i}}}\sum\limits_{\upsilon = 0}^{{\upsilon_i} - 1} {{{ g}_i}(\omega _{t,\upsilon}^i) - {{ g}_i}({{\bar \omega }_t})} } ,{{\bar \omega }_t} - {\omega ^*}} \right\rangle  +  {\mathbb{E}_{t - \tau }}\left\langle {\frac{1}{N}\sum\limits_i {\frac{{1}}{{{\upsilon_i}}}\sum\limits_{j = 1}^{{\upsilon_i}} {g({{\bar \omega }_t}),{{\bar \omega }_t} - {\omega ^*}} } } \right\rangle \nonumber\\
 \le&  \frac{1}{N}\sum\limits_i {\frac{1}{{{\upsilon_i}}}} \sum\limits_{\upsilon = 0}^{{\upsilon_i} - 1} {\left( {\frac{1}{{2{\xi _2} }}{\mathbb{E}_{t - \tau }}\left\| {{{ g}_i}(\omega _{t,\upsilon}^i) - {{ g}_i}({{\bar \omega }_t})} \right\|_2^2 + \frac{{\xi _2} }{2}{\mathbb{E}_{t - \tau }}\left\| {{{\bar \omega }_t} - {\omega ^*}} \right\|_2^2} \right)} \nonumber \\
& +  {\mathbb{E}_{t - \tau }}\left\langle {g({{\bar \omega }_t}),{{\bar \omega }_t} - {\omega ^*}} \right\rangle \label{w1} \\
 \le& \frac{2}{\xi _2 } \cdot \frac{1}{N}\sum\limits_i {\frac{1}{{{\upsilon_i}}}} \sum\limits_{\upsilon = 0}^{{\upsilon_i} - 1} {{\mathbb{E}_{t - \tau }}\left\| {\omega _{t,\upsilon}^i - {{\bar \omega }_t}} \right\|_2^2}  + \frac{\xi _2 }{2}{\mathbb{E}_{t - \tau }}\left\| {{{\bar \omega }_t} - {\omega ^*}} \right\|_2^2 + {\mathbb{E}_{t - \tau }}\left\langle {g({{\bar \omega }_t}),{{\bar \omega }_t} - {\omega ^*}} \right\rangle \label{w2}
\end{align}
where (\ref{w1}) follows from Young's inequality, ${\xi_2}$ is a positive number and (\ref{w2}) follows from Assumption \ref{liang}.

For the third term $W_2$ in (\ref{Per_mar}), we have:
\begin{align}
{W_2} &= {\mathbb{E}_{t - \tau }}\left\| {\frac{1}{N}\sum\limits_i {(d_{t}^i - f_{t}^i) + f_{t}^i} } \right\|_2^2\nonumber\\
&\le 2\underbrace {{\mathbb{E}_{t - \tau }}\left\| {\frac{1}{N}\sum\limits_i {\frac{{1}}{{{\upsilon_i}}}\sum\limits_{k = 0}^{{\upsilon_i-1}} {\left( {{\hat A_i}(O_{t,\upsilon}^i) - {{ A}_i}} \right)(\omega _{t,\upsilon}^i - \omega _i^*) + {\hat Z_i}(O_{t,\upsilon}^i)} } } \right\|_2^2}_{{W_{21}}} + 2\underbrace {{\mathbb{E}_{t - \tau }}\left\| {\frac{1}{N}\sum\limits_i {f_{t}^i} } \right\|_2^2}_{{W_{22}}}
\end{align}

Then we can bound ${W_{21}}$,
\begin{align}
{W_{21}} \le& 2{\mathbb{E}_{t - \tau }}\left\| {\frac{1}{N}\sum\limits_i {\frac{{1}}{{{\upsilon_i}}}\sum\limits_{\upsilon = 0}^{{\upsilon_i} - 1} {\left( {{\hat A_i}(O_{t,\upsilon}^i) - {{ A}_i}} \right)(\omega _{t,\upsilon}^i - \omega _i^*)} } } \right\|_2^2 \nonumber\\
&+ 2{\mathbb{E}_{t - \tau }}\left\| {\frac{1}{N}\sum\limits_i {\frac{{1}}{{{\upsilon_i}}}\sum\limits_{\upsilon = 0}^{{\upsilon_i} - 1} {{\hat Z_i}(O_{t,\upsilon}^i)} } } \right\|_2^2\nonumber\\
 \le& 8{c^2}{\mathbb{E}_{t - \tau }}\left\| {\frac{1}{N}\sum\limits_i {\frac{{1}}{{{\upsilon_i}}}\sum\limits_{\upsilon = 0}^{{\upsilon_i} - 1} {(\omega _{t,\upsilon}^i - {{\bar \omega }_t} + {{\bar \omega }_t} - \omega _i^*)} } } \right\|_2^2 + 2{\mathbb{E}_{t - \tau }}\left\| {\frac{1}{N}\sum\limits_i {\frac{{1}}{{{\upsilon_i}}}\sum\limits_{\upsilon = 0}^{{\upsilon_i} - 1} {{\hat Z_i}(O_{t,\upsilon}^i)} } } \right\|_2^2 \label{c}\\
 \le& 16{c^2}{\mathbb{E}_{t - \tau }}{\Omega _t} + \frac{16{c^2}}{N}{\mathbb{E}_{t - \tau }}\sum\limits_i {\left\| {{{\bar \omega }_t} - \omega _i^*} \right\|_2^2}  + 2{\mathbb{E}_{t - \tau }}\left\| {\frac{1}{N}\sum\limits_i {\frac{{1}}{{{\upsilon_i}}}\sum\limits_{\upsilon = 0}^{{\upsilon_i} - 1} {{\hat Z_i}(O_{t,\upsilon}^i)} } } \right\|_2^2\nonumber\\
 \le& 16{c^2}{\mathbb{E}_{t - \tau }}{\Omega _t} + 16{c^2}{H^2} + 2{\mathbb{E}_{t - \tau }}\left\| {\frac{1}{N}\sum\limits_i {\frac{{1}}{{{\upsilon_i}}}\sum\limits_{\upsilon = 0}^{{\upsilon_i} - 1} {{\hat Z_i}(O_{t,\upsilon}^i)} } } \right\|_2^2 \label{47}
\end{align}
where ${\Omega _t} = \frac{1}{N}\sum\limits_i {\frac{{1}}{{{\upsilon_i}}}\sum\limits_{k = 0}^{{\upsilon_i-1}} {\left\| {\omega _{t,\upsilon}^i - {{\bar \omega }_t}} \right\|_2^2} } $, and (\ref{c}) follows from Lemma \ref{lemma5}.

Then for $W_{22}$,
\begin{align}
{W_{22}} &= {\mathbb{E}_{t - \tau }}\left\| {\frac{1}{N}\sum\limits_i {\frac{{1}}{{{\upsilon_i}}}\sum\limits_{k = 0}^{{\upsilon_i-1}} {\left( {{{ g}_i}(\omega _{t,\upsilon}^i) - {{ g}_i}({{\bar \omega }_t}) + {{ g}_i}({{\bar \omega }_t}) - g({{\bar \omega }_t}) - g({{\bar \omega }_t})} \right)} } } \right\|_2^2 \nonumber\\
&\le 2{\mathbb{E}_{t - \tau }}\left\| {\frac{1}{N}\sum\limits_i {\frac{{1}}{{{\upsilon_i}}}\sum\limits_{k = 0}^{{\upsilon_i-1}} {{{ g}_i}(\omega _{t,\upsilon}^i) - {{ g}_i}({{\bar \omega }_t})} } } \right\|_2^2 + 2{\mathbb{E}_{t - \tau }}\left\| {g({{\bar \omega }_t})} \right\|_2^2 \label{convexcom}\\
&\le \frac{2}{N}\sum\limits_i {\frac{1}{{{\upsilon_i}}}\sum\limits_{\upsilon = 0}^{{\upsilon_i} - 1} {{\mathbb{E}_{t - \tau }}\left\| {{{ g}_i}(\omega _{t,\upsilon}^i) - {{ g}_i}({{\bar \omega }_t})} \right\|_2^2} } + 2{\mathbb{E}_{t - \tau }}\left\| {g({{\bar \omega }_t})} \right\|_2^2\nonumber\\
&\le 8{\Omega _t}+ 2{\mathbb{E}_{t - \tau }}\left\| {g({{\bar \omega }_t})} \right\|_2^2 \label{59}
\end{align}
where (\ref{convexcom}) follows from the fact that $\sum\limits_i {{{\left\| {\frac{1}{N}{z_i}} \right\|}^2}}  \le \frac{1}{N}\sum\limits_i {{{\left\| {{z_i}} \right\|}^2}} $, and (\ref{59}) follows from Assumption \ref{liang}.

By combining the terms $W_{21}$ and $W_{22}$, we can get
\begin{align}
{W_2} \le& 16(2{c^2} + 1){\Omega _t} + 32{c^2}{\left( {{\chi ^2}\left\| {{{\bar \omega }_t} - {\omega ^*}} \right\|_2^2 + {\kappa ^2}} \right)}\nonumber\\
& + 4{\mathbb{E}_{t - \tau }}\left\| {\frac{1}{N}\sum\limits_i {\frac{{1}}{{{\upsilon_i}}}\sum\limits_{\upsilon = 0}^{{\upsilon_i} - 1} {{\hat Z_i}(O_{t,\upsilon}^i)} } } \right\|_2^2 + 4{\mathbb{E}_{t - \tau }}\left\| {g({{\bar \omega }_t})} \right\|_2^2 \label{61}
\end{align}

Next, we study the gradient bias term $W_3$ which plays a central role in Markovian setting. Similar to the analysis of this bias due to Markovian noise in \cite{fedtd,fedhertd}, we rewrite it as
\begin{align*}
{W_3} = &{\mathbb{E}_{t - \tau}}\left\langle {\frac{1}{N}\sum\limits_i {d_{t}^i - f_{t}^i} ,{{\bar \omega }_t} - {\omega ^*}} \right\rangle \\
 =& \underbrace {{\mathbb{E}_{t - \tau}}\left\langle {\frac{1}{N}\sum\limits_i {\frac{{1}}{{{\upsilon_i}}}\sum\limits_{k = 0}^{{\upsilon_i-1}} {{\hat g_i}(\omega _{t,\upsilon}^i) - {{ g}_i}(\omega _{t,\upsilon}^i)} } ,{{\bar \omega }_t} - {{\bar \omega }_{t - \tau}}} \right\rangle }_{{W_{31}}}\\
& + \underbrace {{\mathbb{E}_{t - \tau}}\left\langle {\frac{1}{N}\sum\limits_i {\frac{{1}}{{{\upsilon_i}}}\sum\limits_{k = 0}^{{\upsilon_i-1}} {{\hat g_i}(\omega _{t,\upsilon}^i) - {\hat g_i}(\omega _{t - \tau,k}^i) + {{ g}_i}(\omega _{t - \tau,k}^i) - {{ g}_i}(\omega _{t,\upsilon}^i)} } ,{{\bar \omega }_{t - \tau}} - {\omega ^*}} \right\rangle }_{{W_{32}}}\\
& + \underbrace {{\mathbb{E}_{t - \tau}}\left\langle {\frac{1}{N}\sum\limits_i {\frac{{1}}{{{\upsilon_i}}}\sum\limits_{k = 0}^{{\upsilon_i-1}} {{\hat g_i}(\omega _{t - \tau,k}^i) - {{ g}_i}(\omega _{t - \tau,k}^i)} } ,{{\bar \omega }_{t - \tau}} - {\omega ^*}} \right\rangle }_{{W_{33}}}
\end{align*}

For $W_{31}$, we can conclude that
\begin{align*}
W_{31} &= {\mathbb{E}_{t - \tau}}\left\langle {\frac{1}{N}\sum\limits_i {\frac{{1}}{{{\upsilon_i}}}\sum\limits_{\upsilon = 0}^{{\upsilon_i} - 1} {{\hat g_i}(\omega _{t,\upsilon}^i) - {{ g}_i}(\omega _{t,\upsilon}^i)} } ,{{\bar \omega }_t} - {{\bar \omega }_{t - \tau}}} \right\rangle \nonumber\\
& \le {\mathbb{E}_{t - \tau}}\left[ {{{\left\| {\frac{1}{N}\sum\limits_i {\frac{{1}}{{{\upsilon_i}}}\sum\limits_{\upsilon = 0}^{{\upsilon_i} - 1} {{\hat g_i}(\omega _{t,\upsilon}^i) - {{ g}_i}(\omega _{t,\upsilon}^i)} } } \right\|}_2} \cdot {{\left\| {{{\bar \omega }_t} - {{\bar \omega }_{t - \tau}}} \right\|}_2}} \right]\nonumber\\
& \le {\mathbb{E}_{t - \tau}}\left[ {\frac{\xi _3}{2}\left\| {\frac{1}{N}\sum\limits_i {\frac{{1}}{{{\upsilon_i}}}\sum\limits_{\upsilon = 0}^{{\upsilon_i} - 1} {{\hat g_i}(\omega _{t,\upsilon}^i) - {{ g}_i}(\omega _{t,\upsilon}^i)} } } \right\|_2^2 + \frac{1}{{2{\xi _3}}}\left\| {{{\bar \omega }_t} - {{\bar \omega }_{t - \tau}}} \right\|_2^2} \right]
\end{align*}
where ${\xi _3}$ can be any real positive number. From (\ref{47}), we have:
\begin{align*}
{W_{31}}\le &8{c^2}{\xi _3}{\Omega _t} + 8{\chi ^2}{c^2}{\xi _3}{\mathbb{E}_{t - \tau}}\left\| {{{\bar \omega }_t} - {\omega ^*}} \right\|_2^2 + 8{c^2}{\kappa ^2}{\xi _3} \\
&+ {\xi _3}{\mathbb{E}_{t - \tau}}\left\| {\frac{1}{N}\sum\limits_i {\frac{1}{{{\upsilon_i}}}\sum\limits_{\upsilon = 0}^{{\upsilon_i} - 1} {{\hat Z_i}(O_{t,\upsilon}^i)} } } \right\|_2^2 + \frac{1}{{2{\xi _3}}}\left\| {{{\bar \omega }_t} - {{\bar \omega }_{t - \tau }}} \right\|_2^2
\end{align*}

For $W_{32}$, we have
\begin{align}
{W_{32}} =& {\mathbb{E}_{t - \tau}}\left\langle {\frac{1}{N}\sum\limits_i {\frac{{1}}{{{\upsilon_i}}}\sum\limits_{k = 0}^{{\upsilon_i-1}} {{\hat g_i}(\omega _{t,\upsilon}^i) - {g_i}(\omega _{t - \tau,k}^i) + {{ g}_i}(\omega _{t - \tau,k}^i) - {{ g}_i}(\omega _{t,\upsilon}^i)} } ,{{\bar \omega }_{t - \tau}} - {\omega ^*}} \right\rangle \nonumber\\
\le& {\mathbb{E}_{t - \tau }}{\left\| {\frac{1}{N}\sum\limits_i {\frac{1}{{{\upsilon_i}}}\sum\limits_{\upsilon = 0}^{{\upsilon_i} - 1} {{\hat g_i}(\omega _{t,\upsilon}^i) - {g_i}(\omega _{t - \tau ,k}^i) + {{ g}_i}(\omega _{t - \tau ,k}^i) - {{ g}_i}(\omega _{t,\upsilon}^i)} } } \right\|_2}{\left\| {{{\bar \omega }_{t - \tau }} - {\omega ^*}} \right\|_2}\nonumber\\
\le& {\mathbb{E}_{t - \tau }}\left[ {\frac{1}{N}\sum\limits_i {\frac{{1}}{{{\upsilon_i}}}\sum\limits_{k = 0}^{{\upsilon_i-1}} {{{\left\| {{\hat g_i}(\omega _{t,\upsilon}^i) - {g_i}(\omega _{t - \tau,k}^i)} \right\|}_2} + {{\left\| {{{ g}_i}(\omega _{t - \tau,k}^i) - {{ g}_i}(\omega _{t,\upsilon}^i)} \right\|}_2}} } } \right]{\left\| {{{\bar \omega }_{t - \tau}} - {\omega ^*}} \right\|_2}\nonumber\\
\le& 4{\mathbb{E}_{t - \tau }}\left[ {\frac{1}{N}\sum\limits_i {\frac{{1}}{{{\upsilon_i}}}\sum\limits_{k = 0}^{{\upsilon_i-1}} {{{\left\| {\omega _{t,\upsilon}^i - \omega _{t - \tau,k}^i} \right\|}_2}} } } \right]{\left\| {{{\bar \omega }_{t - \tau}} - {\omega ^*}} \right\|_2} \label{smo}\\
\le& 4{\mathbb{E}_{t - \tau }}\left[ {\frac{1}{N}\sum\limits_i {\frac{{1}}{{{\upsilon_i}}}\sum\limits_{k = 0}^{{\upsilon_i-1}} {{{\left\| {\omega _{t,\upsilon}^i - {{\bar \omega }_t}} \right\|}_2}} }  + \frac{1}{N}\sum\limits_i {\frac{{1}}{{{\upsilon_i}}}\sum\limits_{k = 0}^{{\upsilon_i-1}} {{{\left\| {{{\bar \omega }_t} - {{\bar \omega }_{t - \tau}}} \right\|}_2}} } } \right.\nonumber\\
&\left. { + \frac{1}{N}\sum\limits_i {\frac{{1}}{{{\upsilon_i}}}\sum\limits_{k = 0}^{{\upsilon_i-1}} {{{\left\| {{{\bar \omega }_{t - \tau}} - \omega _{t - \tau,k}^i} \right\|}_2}} } } \right]\left[ {{{\left\| {{{\bar \omega }_{t - \tau}} - {{\bar \omega }_t}} \right\|}_2} + {{\left\| {{{\bar \omega }_t} - {\omega ^*}} \right\|}_2}} \right] \nonumber\\
 = &4{\mathbb{E}_{t - \tau }}\left[ {{\Delta _t}{{\left\| {{{\bar \omega }_{t - \tau}} - {{\bar \omega }_t}} \right\|}_2} + \left\| {{{\bar \omega }_t} - {{\bar \omega }_{t - \tau}}} \right\|_2^2 + {\Delta _{t - \tau}}{{\left\| {{{\bar \omega }_{t - \tau}} - {{\bar \omega }_t}} \right\|}_2}} \right.\nonumber\\
&\left. { + {\Delta _t}{{\left\| {{{\bar \omega }_t} - {\omega ^*}} \right\|}_2} + {{\left\| {{{\bar \omega }_t} - {{\bar \omega }_{t - \tau}}} \right\|}_2}{{\left\| {{{\bar \omega }_t} - {\omega ^*}} \right\|}_2} + {\Delta _{t - \tau}}{{\left\| {{{\bar \omega }_t} - {\omega ^*}} \right\|}_2}} \right]\nonumber\\
 \le& 2{\mathbb{E}_{t - \tau }}\left[ {\frac{2}{{{\xi _4}}}\Delta _t^2 + \frac{2}{{{\xi _4}}}\Delta _{t - \tau}^2 + (2{\xi _4} + \frac{1}{{{\xi _4}}} + 2)\left\| {{{\bar \omega }_{t - \tau}} - {{\bar \omega }_t}} \right\|_2^2 + 3{\xi _4}\left\| {{{\bar \omega }_t} - {\omega ^*}} \right\|_2^2} \right]\label{53}\\
 \le& 2{\mathbb{E}_{t - \tau }}\left[ {\frac{2}{{{\xi _4}}}{\Omega _t} + \frac{2}{{{\xi _4}}}{\Omega _{t - \tau}} + (2{\xi _4} + \frac{1}{{{\xi _4}}} + 2)\left\| {{{\bar \omega }_{t - \tau}} - {{\bar \omega }_t}} \right\|_2^2 + 3{\xi _4}\left\| {{{\bar \omega }_t} - {\omega ^*}} \right\|_2^2} \right] \label{54}
\end{align}
where ${\xi _4}$ can be any positive real number and ${\Delta _t} = \frac{1}{N}\sum\limits_i {\frac{{1}}{{{\upsilon_i}}}\sum\limits_{k = 0}^{{\upsilon_i-1}} {{{\left\| {\omega _{t,\upsilon}^i - {{\bar \omega }_t}} \right\|}_2}} }$; ${\Omega _t} = \sum\limits_i {\frac{{1}}{{{\upsilon_i}}}\sum\limits_{k = 0}^{{\upsilon_i-1}} {\left\| {\omega _{t,\upsilon}^i - {{\bar \omega }_t}} \right\|_2^2} } $; (\ref{smo}) follows from the Lipschitz property and the triangle inequality; (\ref{53}) follows from Young's inequality; (\ref{54}) is derived from the convexity of a square function as follows.
\begin{equation}
{\mathbb{E}_{t - \tau }}\left\| {\frac{1}{N}\sum\limits_i {\frac{{1}}{{{\upsilon_i}}}\sum\limits_{k = 0}^{{\upsilon_i-1}} {\omega _{t,\upsilon}^i - {{\bar \omega }_t}} } } \right\|_2^2 \le \frac{1}{N}\sum\limits_i {\frac{{1}}{{{\upsilon_i}}}\sum\limits_{k = 0}^{{\upsilon_i-1}} {{\mathbb{E}_{t - \tau }}\left\| {\omega _{t,\upsilon}^i - {{\bar \omega }_t}} \right\|_2^2} } 
\end{equation}

For $W_{33}$, we have
\begin{align}
{W_{33}} &= {\mathbb{E}_{t - \tau}}\left\langle {\frac{1}{N}\sum\limits_i {\frac{{1}}{{{\upsilon_i}}}\sum\limits_{k = 0}^{{\upsilon_i-1}} {{\hat g_i}(\omega _{t - \tau,k}^i) - {{ g}_i}(\omega _{t - \tau,k}^i)} } ,{{\bar \omega }_{t - \tau}} - {\omega ^*}} \right\rangle \nonumber\\
& = \frac{1}{N}{\mathbb{E}_{t - \tau}}\sum\limits_i {\frac{{1}}{{{\upsilon_i}}}\sum\limits_{k = 0}^{{\upsilon_i-1}} {{{\left\| {{\hat g_i}(\omega _{t - \tau,k}^i) - {{ g}_i}(\omega _{t - \tau,k}^i)} \right\|}_2}{{\left\| {{{\bar \omega }_{t - \tau }} - {\omega ^*}} \right\|}_2}} }. \nonumber
\end{align}

Then we can further bound it by Lemma \ref{lemma6} as
\begin{align}
{W_{33}} &\le \frac{1}{N}{\mathbb{E}_{t - \tau}}\sum\limits_i {\frac{{1}}{{{\upsilon_i}}}\sum\limits_{k = 0}^{{\upsilon_i-1}} {\left( {{{\left\| {({\hat A_i}(O_{t,k}^i) - {{ A}_i})(\omega _{t - \tau,k}^i - \omega _i^*)} \right\|}_2} + {{\left\| {{Z_i}(O_{t,k}^i)} \right\|}_2}} \right){{\left\| {{{\bar \omega }_{t - \tau }} - {\omega ^*}} \right\|}_2}} } \nonumber\\
& \le \frac{1}{N}{\mathbb{E}_{t - \tau }}\sum\limits_i {\frac{{1}}{{{\upsilon_i}}}\sum\limits_{\upsilon = 0}^{{\upsilon_i} - 1} {\left( {c'{\rho ^{\tau \upsilon_i + k}}{{\left\| {(\omega _{t - \tau ,k}^i - \omega _i^*)} \right\|}_2} + q'{\rho ^{\tau \upsilon_i + k}}} \right){{\left\| {{{\bar \omega }_{t - \tau }} - {\omega ^*}} \right\|}_2}} } \nonumber \\
& \le \frac{ \beta_t^2}{N} {\mathbb{E}_{t - \tau}}\sum\limits_i {\frac{{1}}{{{\upsilon_i}}}\sum\limits_{k = 0}^{{\upsilon_i-1}} {\left( {c'{{\left\| {(\omega _{t - \tau,k}^i - \omega _i^*)} \right\|}_2} + q'} \right){{\left\| {{{\bar \omega }_{t - \tau }} - {\omega ^*}} \right\|}_2}} } \label{56} \\
& \le\frac{ \beta_t^2}{N} {\mathbb{E}_{t - \tau}}\sum\limits_i {\frac{{1}}{{{\upsilon_i}}}\sum\limits_{\upsilon = 0}^{{\upsilon_i} - 1} {\left( {c'{{\left\| {\omega _{t - \tau,k}^i - {{\bar \omega }_{t - \tau}}} \right\|}_2} + c'{{\left\| {{{\bar \omega }_{t - \tau}} - {{\bar \omega }_t}} \right\|}_2} + c'{{\left\| {{{\bar \omega }_t} - \omega _i^*} \right\|}_2} + q'} \right)H} } \nonumber \\
& \le  \beta_t^2 \left[ {\frac{1}{2}c'{\Omega _{t - \tau }} + 2c'{H^2} + \frac{1}{2}c'\left( {{{\chi ^2}}{\mathbb{E}_{t - \tau}}{\left\| {{{\bar \omega }_t} - {\omega ^*}} \right\|_2^2} + {\kappa ^2}} \right) + q'H} \right]\nonumber
\end{align}
where (\ref{56}) is derived from the fact that ${\rho ^{\tau \upsilon_i }} \le \beta_t^2$ for all $i$.

\subsection{Drift term ${\Omega _t}$ }

For the drift term ${\Omega _t}$, we begin with
\begin{align}
{\mathbb{E}}\left\| {\omega _{t,\upsilon}^i - {{\bar \omega }_t}} \right\|_2^2 &= {\beta_t ^2}{\mathbb{E}}\left\| {\sum\limits_{s = 0}^{\upsilon - 1} {{\hat g_i}(\omega _{t,s}^i)} } \right\|_2^2 \nonumber\\
& = {\beta_t ^2} {\mathbb{E}}\left\| {\sum\limits_{s = 0}^{\upsilon - 1} {({\hat A_i}(O_{t,s}^i)(\omega _{t,s}^i - \omega _i^*) + {\hat Z_i}(O_{t,s}^i))} } \right\|_2^2\nonumber\\
& \le 2{\beta_t ^2}{\mathbb{E}}\left\| {\sum\limits_{s = 0}^{\upsilon - 1} {{\hat A_i}(O_{t,s}^i)(\omega _{t,s}^i - \omega _i^*)} } \right\|_2^2 + 2{\beta_t ^2}{\mathbb{E}}\left\| {\sum\limits_{s = 0}^{\upsilon - 1} {{\hat Z_i}(O_{t,s}^i)} } \right\|_2^2\nonumber\\
& \le 2{\beta_t ^2}{c^2}{\mathbb{E}}\left\| {\sum\limits_{s = 0}^{\upsilon - 1} {(\omega _{t,s}^i - \omega _i^*)} } \right\|_2^2 + 2{\beta_t ^2}{\mathbb{E}}\left\| {\sum\limits_{s = 0}^{\upsilon - 1} {{\hat Z_i}(O_{t,s}^i)} } \right\|_2^2 \label{67}\\
& \le 2{\beta_t ^2}{c^2}\upsilon\sum\limits_{s = 0}^{\upsilon - 1} {{\mathbb{E}}\left\| {\omega _{t,s}^i - \omega _i^*} \right\|_2^2}  + 2{\beta_t ^2}{\upsilon^2}{q^2} \label{68}\\
& \le 4{\beta_t ^2}{c^2}\upsilon\sum\limits_{s = 0}^{\upsilon - 1} {\mathbb{E}\left\| {\omega _{t,s}^i - {{\bar \omega }_t}} \right\|_2^2}  + 4{\beta_t ^2}{c^2}{\upsilon^2} {\mathbb{E}\left\| {{{\bar \omega }_t} - \omega _i^*} \right\|_2^2}  + 2{\beta_t ^2}{\upsilon^2}{q^2} \nonumber
\end{align}
where (\ref{67}) and (\ref{68}) are derived from Lemma \ref{lemma5}. Similar to the analysis in the IID setting, we have
\begin{align*}
\frac{1}{{{\upsilon_i}}}\sum\limits_{\upsilon = 0}^{{\upsilon_i} - 1} {{\mathbb{E}}\left\| {\omega _{t,\upsilon}^i - {{\bar \omega }_t}} \right\|_2^2} & \le 2{\beta_t ^2}{c^2}({\upsilon_i} - 1)\sum\limits_{\upsilon = 0}^{{\upsilon_i} - 1} {{\mathbb{E}}\left\| {\omega _{t,\upsilon}^i - {{\bar \omega }_t}} \right\|_2^2} \\
& +  \frac{{2{\beta_t ^2}{c^2}({\upsilon_i} - 1)(2{\upsilon_i} - 1)}}{3}\mathbb{E}\left\| {{{\bar \omega }_t} - \omega _i^*} \right\|_2^2\\
& + \frac{{{\beta_t ^2}{q^2}({\upsilon_i} - 1)(2{\upsilon_i} - 1)}}{3}
\end{align*}

After minor arrangements, we have
\begin{align*}
\frac{1}{{{\upsilon_i}}}\sum\limits_{\upsilon = 0}^{{\upsilon_i} - 1} {{\mathbb{E}}\left\| {\omega _{t,\upsilon}^i - {{\bar \omega }_t}} \right\|_2^2} & \le \frac{{2{\beta_t ^2}{c^2}({\upsilon_i} - 1)(2{\upsilon_i} - 1)}}{{3(1 - 2{\beta_t ^2}{c^2}{\upsilon_i}({\upsilon_i} - 1))}}{\mathbb{E}}\left\| {{{\bar \omega }_t} - \omega _i^*} \right\|_2^2\\
& + \frac{{{\beta_t ^2}{q^2}({\upsilon_i} - 1)(2{\upsilon_i} - 1)}}{{3(1 - 2{\beta_t ^2}{c^2}{\upsilon_i}({\upsilon_i} - 1))}}
\end{align*}

Define $C' = \mathop {\max }\limits_i \{ 2{\beta_t ^2}{c^2}{\upsilon_i}({\upsilon_i} - 1)\} $, then we have
\begin{align}
\frac{1}{{{\upsilon_i}}}\sum\limits_{\upsilon = 0}^{{\upsilon_i} - 1} {{\mathbb{E}}\left\| {\omega _{t,\upsilon}^i - {{\bar \omega }_t}} \right\|_2^2} & \le \frac{{2{\beta_t ^2}{c^2}({\upsilon_i} - 1)(2{\upsilon_i} - 1)}}{{3(1 - C')}}{\mathbb{E}}\left\| {{{\bar \omega }_t} - \omega _i^*} \right\|_2^2\nonumber\\
& + \frac{{{\beta_t ^2}{q^2}({\upsilon_i} - 1)(2{\upsilon_i} - 1)}}{{3(1 - C')}}
\end{align}

Then, the drift term can be represented as 
\begin{align} \label{driftterm}
&\frac{1}{N}\sum\limits_i {\frac{{1}}{{{\upsilon_i}}}\sum\limits_{\upsilon = 0}^{{\upsilon_i} - 1} {\mathbb{E}\left\| {\omega _{t,\upsilon}^i - {{\bar \omega }_t}} \right\|_2^2} } \nonumber \\
 \le& \frac{{2{\beta_t ^2}{c^2}({\upsilon_{\max}} - 1)(2{\upsilon_{\max}} - 1)}}{{3(1 - C')N}}\sum\limits_i {\mathbb{E}\left\| {{{\bar \omega }_t} - \omega _i^*} \right\|_2^2}  + \frac{{{\beta_t ^2}{q^2}({\upsilon_{\max}} - 1)(2{\upsilon_{\max}} - 1)}}{{3(1 - C')}}\nonumber\\
 \le & \frac{{2{\beta_t ^2}{c^2}({\upsilon_{\max}} - 1)(2{\upsilon_{\max}} - 1)}}{{3(1 - C')}}\left( {{\chi ^2}\mathbb{E}\left\| {{{\bar \omega }_t} - {\omega ^*}} \right\|_2^2 + {\kappa ^2}} \right)\nonumber\\
& + \frac{{{\beta_t ^2}{q^2}({\upsilon_{\max}} - 1)(2{\upsilon_{\max}} - 1)}}{{3(1 - C')}}
\end{align}

Next, we bound $\mathbb{E}{\left\| {{{\bar \omega }_t} - {{\bar \omega }_{t - \tau }}} \right\|^2}$ since it appears in $W_3$.

\begin{align}
&\left\| {{{\bar \omega }_{l + 1}} - {{\bar \omega }_l}} \right\|_2^2 = {\beta_t ^2}\left\| {\frac{1}{N}\sum\limits_i {\frac{{1}}{{{\upsilon_i}}}\sum\limits_{\upsilon = 0}^{{\upsilon_i} - 1} {{\hat g_i}(\omega _{l,k}^i)} } } \right\|_2^2 \nonumber\\
= &{\beta_t ^2}\left\| {\frac{1}{N}\sum\limits_i {\frac{{1}}{{{\upsilon_i}}}\sum\limits_{\upsilon = 0}^{{\upsilon_i} - 1} {({\hat A_i}(O_{l,k}^i)(\omega _{l,k}^i - \omega _i^*) + {\hat Z_i}(O_{l,k}^i))} } } \right\|_2^2\nonumber\\
 \le& 2c{\beta_t ^2}\left\| {\frac{1}{N}\sum\limits_i {\frac{{1}}{{{\upsilon_i}}}\sum\limits_{\upsilon = 0}^{{\upsilon_i} - 1} {\omega _{l,k}^i - \omega _i^*} } } \right\|_2^2 + 2{\beta_t ^2}\left\| {\frac{1}{N}\sum\limits_i {\frac{{1}}{{{\upsilon_i}}}\sum\limits_{\upsilon = 0}^{{\upsilon_i} - 1} {{\hat Z_i}(O_{l,k}^i)} } } \right\|_2^2\nonumber\\
 \le &4c{\beta_t ^2}\left\| {\frac{1}{N}\sum\limits_i {\frac{{1}}{{{\upsilon_i}}}\sum\limits_{\upsilon = 0}^{{\upsilon_i} - 1} {(\omega _{l,k}^i - {{\bar \omega }_t})} } } \right\|_2^2 + 4c{\beta_t ^2}\left\| {\frac{1}{N}\sum\limits_i {\frac{{1}}{{{\upsilon_i}}}\sum\limits_{\upsilon = 0}^{{\upsilon_i} - 1} {({{\bar \omega }_t} - \omega _i^*)} } } \right\|_2^2\nonumber\\
& + 2{\beta_t ^2}\left\| {\frac{1}{N}\sum\limits_i {\frac{{1}}{{{\upsilon_i}}}\sum\limits_{\upsilon = 0}^{{\upsilon_i} - 1} {{\hat Z_i}(O_{l,k}^i)} } } \right\|_2^2\nonumber\\
\le &4c{\beta_t ^2}\left\| {\frac{1}{N}\sum\limits_i {\frac{{1}}{{{\upsilon_i}}}\sum\limits_{\upsilon = 0}^{{\upsilon_i} - 1} {(\omega _{l,k}^i - {{\bar \omega }_t})} } } \right\|_2^2 + 4c{\beta_t ^2}\left( {{\chi ^2}\left\| {{{\bar \omega }_t} - {\omega ^*}} \right\|_2^2 + {\kappa ^2}} \right)\nonumber\\
& + 2{\beta_t ^2}\left\| {\frac{1}{N}\sum\limits_i {\frac{{1}}{{{\upsilon_i}}}\sum\limits_{\upsilon = 0}^{{\upsilon_i} - 1} {{\hat Z_i}(O_{l,k}^i)} } } \right\|_2^2 \label{71}
\end{align}

Then we can establish the bound in conditional expectation on $t-2\tau$:
\begin{align}
&{\mathbb{E}_{t - 2\tau}}\left\| {{{\bar \omega }_t} - {{\bar \omega }_{t - \tau}}} \right\|_2^2\nonumber\\
\le &\tau\sum\limits_{s = t - \tau}^{t - 1} {\left\| {{{\bar \omega }_{s + 1}} - {{\bar \omega }_s}} \right\|_2^2} \nonumber\\
\le &2{\beta_t ^2}\tau\sum\limits_{s = t - \tau}^{t - 1} {{\mathbb{E}_{t - 2\tau}}\left[ {2c\left\| {\frac{1}{N}\sum\limits_i {\frac{{1}}{{{\upsilon_i}}}\sum\limits_{\upsilon = 0}^{{\upsilon_i} - 1} {(\omega _{s,k}^i - {{\bar \omega }_t})} } } \right\|_2^2 + 2c\left( {{\chi ^2}\left\| {{{\bar \omega }_s} - {\omega ^*}} \right\|_2^2 + {\kappa ^2}} \right)} \right.}\nonumber\\
& + \left. {\left\| {\frac{1}{N}\sum\limits_i {\frac{{1}}{{{\upsilon_i}}}\sum\limits_{\upsilon = 0}^{{\upsilon_i} - 1} {{\hat Z_i}(O_{s,k}^i)} } } \right\|_2^2} \right]\label{72}\\
\le& 4{\beta_t ^2}c\tau\sum\limits_{s = 0}^\tau {{\mathbb{E}_{t - 2\tau}}} {\Omega _{t - s}} + 4{\beta_t ^2}c{\tau^2}\left( {{\chi ^2}{H^2} + {\kappa ^2}} \right)\nonumber \\
& + 2{\beta_t ^2}{\tau^2}{\mathbb{E}_{t - 2\tau}}\left[ {\frac{1}{{{N^2}}}\sum\limits_i {\frac{{{{[q']}^2}}}{{{\upsilon_i}}}}  + \frac{{2qq'\rho ({\upsilon_{\max}} - 1)}}{{{N^2}(1 - \rho )}}\sum\limits_i {\frac{1}{{{{{\upsilon _i^2}}}}}} } \right.\nonumber \\
& + \left. { \frac{{2{{[q']}^2}\rho ({\upsilon_{\max}} - 1)}}{{{N^2}}{(1 - \rho) }}\sum\limits_{i < j} {\frac{1}{{{\upsilon_i}{\upsilon_j}}}{\rho ^{^{{2\tau(\upsilon_i+\upsilon_j)}}}}}} \right] \label{73}
\end{align}
where (\ref{72}) follows from (\ref{71}); (\ref{73}) follows from Lemma \ref{lemma5}; (146) follows from that fact that ${\rho ^{{\upsilon_i}\tau}} \le {\rho ^{{{\upsilon_{\min }}}\tau}} \le \beta_t^2$ for all $i$.

Finally, we choose $\tau = \left\lfloor {\frac{{{2{\log }_\rho }\beta_t }}{{{{\upsilon_{\min }}}}}} \right\rfloor$ and by combining $W_1$, $W_2$ and $W_3$, we have:
\begin{align}
&{\mathbb{E}_{t - 2\tau}}\left\| {{{\bar \omega }_{t + 1}} - {\omega ^*}} \right\|_2^2\nonumber\\
\le &{\mathbb{E}_{t - 2\tau}}\left\| {{{\bar \omega }_t} - {\omega ^*}} \right\|_2^2 + 2{\beta _t} {\bar {\upsilon}}{\mathbb{E}_{t - 2\tau}}\left\langle {\frac{1}{N}\sum\limits_i {f_{t}^i} ,{{\bar \omega }_t} - {\omega ^*}} \right\rangle  + {\beta _t ^2}{\bar {\upsilon}}^2{\mathbb{E}_{t - 2\tau}}\left\| {\frac{1}{N}\sum\limits_i {d_{t}^i} } \right\|_2^2\nonumber\\
& + 2\beta _t {\bar {\upsilon}}{\mathbb{E}_{t - 2\tau}}\left\langle {\frac{1}{N}\sum\limits_i {(d_{t}^i - f_{t}^i)} ,{{\bar \omega }_t} - {\omega ^*}} \right\rangle \nonumber\\
\le & 2{\beta _t}{\bar {\upsilon}}\left( {\frac{2}{{{\xi _2}}} \cdot {\mathbb{E}_{t - 2\tau }}{\Omega _t} + \frac{{{\xi _2}}}{2}{\mathbb{E}_{t - 2\tau }}\left\| {{{\bar \omega }_t} - {\omega ^*}} \right\|_2^2 + {\mathbb{E}_{t - 2\tau }}\left\langle {g({{\bar \omega }_t}),{{\bar \omega }_t} - {\omega ^*}} \right\rangle } \right) \nonumber\\
&  + 16\beta _t^2{\bar {\upsilon}}^2\left( {(2{c^2} + 1){\mathbb{E}_{t - 2\tau }}{\Omega _t} + 2{c^2}\left( {{\chi ^2}\mathbb{E}\left\| {{{\bar \omega }_t} - {\omega ^*}} \right\|_2^2 + {\kappa ^2}} \right)} \right) \nonumber\\
& + 4\beta _t^2{\bar {\upsilon}}^2\left( {{\mathbb{E}_{t - 2\tau }}\left\| {\frac{1}{N}\sum\limits_i {\frac{1}{{{\upsilon_i}}}\sum\limits_{\upsilon = 0}^{{\upsilon_i} - 1} {{\hat Z_i}(O_{t,\upsilon}^i)} } } \right\|_2^2 + {\mathbb{E}_{t - 2\tau }}\left\| {g({{\bar \omega }_t})} \right\|_2^2} \right) \nonumber \\
&+ 2{\beta _t}{\bar {\upsilon}}\left( {8{c^2}{\xi _3}{\mathbb{E}_{t - 2\tau }}{\Omega _t} + 8{\chi ^2}{c^2}{\xi _3}{\mathbb{E}_{t - 2\tau }}\left\| {{{\bar \omega }_t} - {\omega ^*}} \right\|_2^2 + 8{c^2}{\kappa ^2}{\xi _3}} \right) \nonumber\\
& + 2{\beta _t}{\bar {\upsilon}}\left( {{\xi _3}{\mathbb{E}_{t - 2\tau }}\left\| {\frac{1}{N}\sum\limits_i {\frac{1}{{{\upsilon_i}}}\sum\limits_{\upsilon = 0}^{{\upsilon_i} - 1} {{\hat Z_i}(O_{t,\upsilon}^i)} } } \right\|_2^2 + \frac{1}{{2{\xi _3}}}{\mathbb{E}_{t - 2\tau }}\left\| {{{\bar \omega }_t} - {{\bar \omega }_{t - \tau }}} \right\|_2^2} \right) \nonumber\\
& + 4{\beta _t}{\bar {\upsilon}}{\mathbb{E}_{t - 2\tau }}\left[ {\frac{2}{{{\xi _4}}}{\Omega _t} + \frac{2}{{{\xi _4}}}{\omega _{t - \tau }} + (2{\xi _4} + \frac{1}{{{\xi _4}}} + 2)\left\| {{{\bar \omega }_{t - \tau }} - {{\bar \omega }_t}} \right\|_2^2 + 3{\xi _4}\left\| {{{\bar \omega }_t} - {\omega ^*}} \right\|_2^2} \right] \nonumber\\
& + 2\beta _t^3{\bar {\upsilon}}\left( {\frac{1}{2}c'{\mathbb{E}_{t - 2\tau }}{\Omega _{t - \tau }} + 3c'{H^2} + q'H} \right) \label{74}
\end{align}
where (\ref{74}) is derived from (\ref{w2}), (\ref{61}), (\ref{54}), and (\ref{47}).

If we choose ${\xi _2} = \frac{{{\lambda}}}{{4}}$, ${\xi _3} = \beta _t$, ${\xi _4} = \frac{{{\lambda}}}{{48}}$, and step size ${\beta _t} \le \frac{\lambda }{{4\left( {32{\bar \upsilon}^2{\chi ^2}{c^2} + 16{\bar \upsilon}^2 + 16{\chi ^2}{c^2} + {\chi ^2}c'} \right)}}$, by combining (\ref{driftterm}) and Lemma \ref{lemma5}, we have
\begin{align*}
{\mathbb{E}_{t - 2\tau }}&\left\| {{{\bar \omega }_{t + 1}} - {\omega ^*}} \right\|_2^2 \le \left( {1 - \frac{{\beta {\bar {\upsilon}}\lambda }}{4}} \right){\mathbb{E}_{t - 2\tau }}\left\| {{{\bar \omega }_t} - {\omega ^*}} \right\|_2^2\\
&+ \left[ {\frac{{400{\beta _t}{\bar {\upsilon}}}}{\lambda } + 16\beta _t^2{\bar {\upsilon}}^2(2{c^2} + 1) + 16\beta _t^2{\bar {\upsilon}}{c^2}} \right]{\mathbb{E}_{t - 2\tau }}{\Omega _t}\\
& + \left( {\frac{{384{\beta _t}{\bar {\upsilon}}}}{\lambda } + \beta _t^2{\bar {\upsilon}}c'} \right){\mathbb{E}_{t - 2\tau }}{\Omega _{t - \tau }}  + \left[ {{\bar {\upsilon}} + 4{\beta _t}{\bar {\upsilon}}\left( {\frac{\lambda }{{24}} + \frac{{48}}{\lambda } + 2} \right)} \right]{\mathbb{E}_{t - 2\tau }}\left\| {{{\bar \omega }_t} - {{\bar \omega }_{t - \tau }}} \right\|_2^2 \\
&+\left( {4\beta _t^2{\bar \upsilon}^2 + 2\beta _t^2{\bar {\upsilon}}} \right){\mathbb{E}_{t - 2\tau }}\left\| {\frac{1}{N}\sum\limits_i {\frac{1}{{{\upsilon_i}}}\sum\limits_{\upsilon = 0}^{{\upsilon_i} - 1} {{\hat Z_i}(O_{t,\upsilon}^i)} } } \right\|_2^2 \\
& + \left( {32\beta _t^2{\bar {\upsilon}}^2{c^2} + \beta _t^2{\bar {\upsilon}}c' + 16\beta _t^2{\bar {\upsilon}}^2c} \right){\kappa ^2} + 2\beta _t^3{\bar {\upsilon}}H\left( {2c'H + q'} \right)    
\end{align*}    

If step sizes satisfy that
\begin{equation*}
{\beta _t} \le \min \left\{ {\frac{2}{{\lambda (2{c^2}{\bar {\upsilon}} + {c^2} + {\bar {\upsilon}})}},\frac{1}{{\lambda c'}},\frac{1}{{4\left( {\frac{\lambda }{{24}} + \frac{{48}}{\lambda } + 2} \right)}}} \right\},
\end{equation*}
then $\frac{{400{\beta _t}{\bar {\upsilon}}}}{\lambda } + 16\beta _t^2{\bar \upsilon}^2(2{c^2} + 1) + 16\beta _t^2{\bar {\upsilon}}{c^2} \le \frac{{432{\beta _t}{\bar {\upsilon}}}}{\lambda }$, $\frac{{384{\beta _t}{\bar {\upsilon}}}}{\lambda } + \beta _t^2{\bar {\upsilon}}c' \le \frac{{385{\beta _t}{\bar {\upsilon}}}}{\lambda }$, and ${\bar {\upsilon}} + 4{\beta _t}{\bar {\upsilon}}\left( {\frac{\lambda }{{24}} + \frac{{48}}{\lambda } + 2} \right) \le 2{\bar {\upsilon}}$ hold. Then (\ref{74}) can be simplified as
\begin{align*}
{\mathbb{E}_{t - 2\tau }}&\left\| {{{\bar \omega }_{t + 1}} - {\omega ^*}} \right\|_2^2 \le \left( {1 - \frac{{\beta_t {\bar {\upsilon}}\lambda }}{4}} \right){\mathbb{E}_{t - 2\tau }}\left\| {{{\bar \omega }_t} - {\omega ^*}} \right\|_2^2\\
+ &\left( {\frac{{817{\beta _t}{\bar {\upsilon}}}}{\lambda } + 8\beta _t^2{\tau ^2}{\bar {\upsilon}}c} \right)\left( {\frac{{2{\chi ^2}{c^2}({\upsilon_{\max}} - 1)(2{\upsilon_{\max}} - 1)}}{{3(1 - C')}}\left( {{\chi ^2}{H^2} + {\kappa ^2}} \right)} \right)\\
&+ \left( {\frac{{817{\beta _t}{\bar {\upsilon}}}}{\lambda } + 8\beta _t^2{\tau ^2}{\bar {\upsilon}}c} \right)\left( {\frac{{{\chi ^2}{q^2}({\upsilon_{\max}} - 1)(2{\upsilon_{\max}} - 1)}}{{3(1 - C')}}} \right) \\
+ &\left( {4\beta _t^2{\bar \upsilon}^2 + 2\beta _t^2{\bar {\upsilon}} + 4\beta _t^2{\tau ^2}{\bar {\upsilon}}} \right)\left( {\frac{1}{{{N^2}}}\sum\limits_i {\frac{{{{[q']}^2}}}{{{\upsilon_i}}}}  + \frac{{2qq'\rho ({\upsilon_{\max}} - 1)}}{{{N^2}(1 - \rho )}}\sum\limits_i {\frac{1}{{{{{\upsilon _i^2}}}}}}  } \right)\\
&+\left( {4\beta _t^2{\bar \upsilon}^2 + 2\beta _t^2{\bar {\upsilon}} + 4\beta _t^2{\tau ^2}{\bar {\upsilon}}} \right)\left(  \frac{{2{{[c']}^2}\rho ({\upsilon_{\max}} - 1)}}{{{N^2}}{(1 - \rho) }}\sum\limits_{i < j} {\frac{1}{{{\upsilon_i}{\upsilon_j}}}{\rho ^{2\tau({\upsilon_i}+{\upsilon_j})}}}\right)\\
+ &\left( {32\beta _t^2{\bar \upsilon}^2{c^2} + \beta _t^2{\bar {\upsilon}}c' + 16\beta _t^2{\bar \upsilon}^2c + 8\beta _t^2{\tau ^2}{\bar {\upsilon}}c} \right){\kappa ^2} + 2\beta _t^3{\bar {\upsilon}}H\left( {2c'H + q' } \right)
\end{align*}

Then let  ${{\hat \upsilon}} = \frac{1}{N}\sum\limits_i {\frac{1}{{{\upsilon_i}}}} $ and $\hat \upsilon^2 = \frac{1}{N}\sum\limits_i {\frac{1}{{{{{\upsilon _i^2}}}}}} $, it can also be shown as 
\begin{align*}
\mathbb{E}\left\| {{{\bar \omega }_t} - {\omega ^*}} \right\|_2^2 \le {e^{ - \frac{{\lambda \beta \bar \nu T}}{4}}}\mathbb{E}\left\| {{{\bar \omega }_0} - {\omega ^*}} \right\|_2^2 + {C_1}{\beta ^3} + {C_2}{\chi ^2} + {C_3}\frac{\beta }{N} + {C_4}\beta 
\end{align*}
where ${C_\lambda } = \frac{1}{{3(1 - C')}}$, ${C_{K,\kappa }} = \frac{{8{\beta_t ^2}{c^2}({\upsilon_{\max }} - 1)(2{\upsilon_{\max }} - 1)}}{{3(1 - C')}}\left( {{\chi ^2}{H^2} + {\kappa ^2}} \right)$, ${C_K} = \frac{{8\left( {2{{\upsilon}_{\max }} + 2{\tau ^2} + 1} \right)}}{\lambda } \cdot \frac{{2{{[c']}^2}\rho ({\upsilon_{\max }} - 1)}}{{(1 - \rho )}}$, and 

\begin{align*}
{C_1} &= \frac{{8{\tau ^2}c{C_\lambda }{C_{K,\kappa }}}}{\lambda },\\
{C_2} &= \frac{{817{C_\lambda }\left( {{C_{K,\kappa }} + {C_K}} \right)}}{{{\lambda ^2}}}+\frac{2 H\left( {2c'H + q' } \right)}{\lambda},\\
{C_3} &= \frac{{8\left( {2{{\upsilon_{\max }}} + 2{\tau ^2} + 1} \right)}}{\lambda }\left( {{{[q']}^2}{{\hat \upsilon}_{\max }} + \frac{{2qq'\rho ({\upsilon_{\max }} - 1){\hat \upsilon _{\max }^2}}}{{(1 - \rho )}}} \right),\\
{C_4} &= \frac{{4\left( {32{c^2}{{\upsilon_{\max }}} + 16{{\upsilon_{\max }}}c + 8{\tau ^2}c + c'} \right){\kappa ^2} }}{\lambda }
\end{align*}

\subsection{Proof of Theorem 1}
Using ${\omega _{k + 1}^*}$ as shorthand notation for ${\omega ^*}({\theta _{k + 1}})$, we start by decomposing the error of the critics as 
\begin{align*}
&{\left\| {{\omega _{k + 1}} - \omega _{k + 1}^*} \right\|^2}\\
= &{\left\| {{\omega _{k + 1}} - \omega _k^* + \omega _k^* - \omega _{k + 1}^*} \right\|^2}\\
= &\underbrace {{{\left\| {{\omega _{k + 1}} - \omega _k^*} \right\|}^2}}_{{J_1}} + 2\underbrace {\left\langle {{\omega _{k + 1}} - \omega _k^*,\omega _k^* - \omega _{k + 1}^*} \right\rangle }_{{J_2}} + \underbrace {{{\left\| {\omega _k^* - \omega _{k + 1}^*} \right\|}^2}}_{{J_3}}
\end{align*}
For $J_1$, since ${\omega _{k + 1}} = {\omega _{k,T}}$, we first analyze from Proposition 1
\begin{align*}
\mathbb{E}\left\| {{{ \omega  }_{k, T}} - \omega _{{k}}^*} \right\|_2^2 &\le \left( {1 - \frac{{\beta \overline \upsilon  \lambda }}{4}} \right)\left\| {{{ \omega  }_{k, T - 1}} - \omega _{{k}}^*} \right\|_2^2 + {C_1}{\beta ^4} + {C_2}{\beta ^3} + {C_3}\frac{{{\beta ^2}}}{N} + {C_4}{\beta ^2}
\end{align*}
Then using induction, we have
\begin{align*}
\mathbb{E}\left\| {{\omega _{k,T}} - \omega _k^*} \right\|_2^2 &\le {\left( {1 - \frac{{\beta \bar \upsilon \lambda }}{4}} \right)^T}\left\| {{\omega _k} - \omega _k^*} \right\|_2^2 + T\left( {{C_1}{\beta ^4} + {C_2}{\beta ^3} + {C_3}\frac{{{\beta ^2}}}{N} + {C_4}{\beta ^2}} \right)
\end{align*}
For $J_2$, we have
\begin{align*}
&\left\langle {{\omega _{k + 1}} - \omega _k^*,\omega _k^* - \omega _{k + 1}^*} \right\rangle \\
 = &\underbrace {\left\langle {{\omega _{k + 1}} - \omega _k^*,\omega _k^* - \omega _{k + 1}^* - {{\left( {\nabla \omega _k^*} \right)}^{\rm{T}}}\left( {{\theta _{k + 1}} - {\theta _k}} \right)} \right\rangle }_{{J_{21}}}  + \underbrace {\left\langle {{\omega _{k + 1}} - \omega _k^*,{{\left( {\nabla \omega _k^*} \right)}^{\rm{T}}}\left( {{\theta _{k + 1}} - {\theta _k}} \right)} \right\rangle }_{{J_{22}}}
\end{align*}
Taking expectation up to round $k+1$, we have:
\begin{align*}
{J_{21}} &\overset{(a)}{\le} \frac{{{L_{\omega\theta}}}}{2}{\mathbb{E}}{\left\| {{\omega _{k + 1}} - \omega _k^*} \right\|_2}\left\| {{\theta _{k + 1}} - {\theta _k}} \right\|_2^2\\
& \le \frac{{{L_{\omega\theta}}\alpha _k^2}}{2}{\mathbb{E}}{\left\| {{\omega _{k + 1}} - \omega _k^*} \right\|_2}\left\| {\frac{1}{N}\sum\limits_i {{{\hat h}_i}({\theta _k},{\omega _{k + 1}})} } \right\|_2^2\\
& \overset{(b)}{\le} \frac{{ {L_{\omega\theta}}}\alpha _k^2C_f^2}{4}{\mathbb{E}}\left\| {{\omega _{k + 1}} - \omega _k^*} \right\|_2^2 + \frac{{{L_{\omega\theta}}\alpha _k^2}}{{4 }}\left\| {\frac{1}{N}\sum\limits_i {{{\hat h}_i}({\theta _k},{\omega _{k + 1}})} } \right\|_2^2
\end{align*}
where (a) is derived from Lemma \ref{nomega} and (b) follows from Young's inequality and the fact that $\left\| {{{\hat h}^i}({\omega _{k + 1}},{\theta _k})} \right\| \le H{C_\psi } = {C_f}$.
\\
$J_{22}$ can be bounded as
\begin{align*}
{J_{22}} &\le \mathbb{E}\left\| {{\omega _{k + 1}} - \omega _k^*} \right\|_2\left\| {{{\left( {\nabla \omega _k^*} \right)}^{\rm{T}}}\left( {{\theta _{k + 1}} - {\theta _k}} \right)} \right\|_2\\
& \overset{(a)}{\le} {\alpha_k}L_\omega^2 {\mathbb{E}}\left\| {{\omega _{k + 1}} - \omega _k^*} \right\|_2^2 + \frac{{\alpha _k}}{{4 }}\mathbb{E}\left\| {\frac{1}{N}\sum\limits_{i = 1}^N {{{\hat h}^i}({{\omega }_{k+1}},{\theta_k})} } \right\|_2^2
\end{align*}
where (a) follows from Lemma \ref{omega} and Young's inequality.
\\
${\mathbb{E}}J_3$ can be bounded as
\begin{align*}
{\mathbb{E}}{J_3} \le L_{\omega}^2{\mathbb{E}}\left\| {{\theta _{k + 1}} - {\theta _k}} \right\|_2^2 \le \alpha _k^2L_{\omega}^2\mathbb{E}\left\| {\frac{1}{N}\sum\limits_{i = 1}^N {{{\hat h}^i}({{\omega }_{k+1}},{\theta_k})} } \right\|_2^2
\end{align*}
Thus, ${\left\| {{\omega _{k + 1}} - \omega _{k + 1}^*} \right\|^2}$ can be bounded as
\begin{align*}
&{\left\| {{\omega _{k + 1}} - \omega _{k + 1}^*} \right\|^2} \le \left( {1 + {\alpha _k}L_\omega ^2 + \frac{{{L_{\omega \theta }}\alpha _k^2C_f^2}}{4}} \right)\left\| {{\omega _{k + 1}} - \omega _k^*} \right\|_2^2 + \left( {\frac{{{L_{\omega \theta }}\alpha _k^2}}{4} + \frac{{{\alpha _k}}}{4} + \alpha _k^2L_\omega ^2} \right)\mathbb{E}\left\| {\frac{1}{N}\sum\limits_{i = 1}^N {{{\hat h}^i}({{\omega }_{k+1}},{\theta_k})} } \right\|_2^2
\end{align*}

Then we move to the outer loop, we have 
\begin{align*}
&J({\theta_{k + 1}}) \ge J({\theta_k}) + \left\langle {\nabla J({\theta_k}),{\theta_{k + 1}} - {\theta_k}} \right\rangle  - \frac{L}{2}\left\| {{\theta_{k + 1}} - {\theta_k}} \right\|_2^2\\
 \ge &J({\theta_k}) - \frac{{{\alpha_k ^2}L}}{2}\left\| {\frac{1}{N}\sum\limits_{i = 1}^N {{{\hat h}^i}({{\omega }_{k+1}},{\theta_k})} } \right\|_2^2 + \alpha_k \left\langle {\nabla J({\theta_k}),\frac{1}{N}\sum\limits_{i = 1}^N {{{\hat h}^i}({{\omega }_{k+1}},{\theta_k})}} \right\rangle \\
 = &J({\theta_k}) + \frac{\alpha_k }{2}\left\| {\nabla J({\theta_k})} \right\|_2^2 + \frac{\alpha_k }{2}(1-{\alpha_k}L)\left\| {\frac{1}{N}\sum\limits_{i = 1}^N {{{\hat h}^i}({{\omega }_{k+1}},{\theta_k})} } \right\|_2^2 - \frac{\alpha_k }{2}\underbrace{\left\| {\frac{1}{N}\sum\limits_{i = 1}^N {{{\hat h}^i}({{\omega }_{k+1}},{\theta_k}) - \nabla J({\theta_k})} } \right\|_2^2}_{A}\\
\end{align*}
where the last inequality is derived using Young's inequality $\left\langle {a,b} \right\rangle  \ge  - {\textstyle{1 \over 2}}{\left\| a \right\|^2} +  - {\textstyle{1 \over 2}}{\left\| b \right\|^2}$.

The gradient bias $A$ can be bounded as 
\begin{align*}
&\left\| {\frac{1}{N}\sum\limits_{i = 1}^N {{{\hat h}^i}({{\omega }_{k+1}},{\theta_k}) - \nabla J({\theta_k})} } \right\|_2^2 \\
 \le &5\underbrace {\left\| {\frac{1}{N}\sum\limits_{i = 1}^N {{{{\hat h}^i}({{\omega }_{k + 1}},{\theta _k}) - {{\hat h}^i}({{\omega }_{k + 1}^*},{\theta _k})}} } \right\|_2^2}_{{A_1}}  + 5\underbrace {\left\| {\frac{1}{N}\sum\limits_{i = 1}^N {{{{\hat h}^i}({{\omega }_{k + 1}^*},{\theta _k}) - {h^i}(\omega _{k + 1}^*,{\theta _k})}} } \right\|_2^2}_{{A_2}}\\
& + 5\underbrace {\left\| {\frac{1}{N}\sum\limits_{i = 1}^N {{{h^i}(\omega _{k + 1}^*,{\theta _k}) - {h^i}(\omega _{k + 1}^{i*},{\theta _k})}} } \right\|_2^2}_{{A_3}} + 5\underbrace {\left\| {\frac{1}{N}\sum\limits_{i = 1}^N {{{h^i}(\omega _{k + 1}^{i*},{\theta _k}) - {h^i}(\omega _k^{i*},{\theta _k})}} } \right\|_2^2}_{{A_4}} + 5\underbrace {\left\| {\frac{1}{N}\sum\limits_{i = 1}^N {{{h^i}(\omega _k^{i*},{\theta _k}) - \nabla {J^i}({\theta _k})}} } \right\|_2^2}_{{A_5}}
\end{align*}
By decomposition and expectation, the gradient bias can be further bounded as
\begin{align}
A \le &\frac{{40{H^2}\left( {1 + (\eta  - 1)\rho } \right)}}{{M(1 - \rho )}}+ \frac{{40{\kappa ^2}}}{{{c^2}}} + 20\alpha _k^2L_{\omega}^2\left\| {\frac{1}{N}\sum\limits_{i = 1}^N {{{\hat h}^i}({{\omega }_{k+1}},{\theta_k})} } \right\|_2^2  + 5{\xi _{critic}}  + 20\left( {C_g^2 + 2\left( {{\chi ^2} + 1} \right)} \right)\left\| {{{\omega }_{k + 1}} - \omega _{k + 1}^*} \right\|_2^2 
\end{align}
Consider the difference of the Lyapunov function ${\mathbb{V}^k}: =  - J({\theta _k}) + \left\| {{{\omega }_k} - \omega _k^*} \right\|_2^2$, given by
\begin{align*}
&\mathbb{E}[{V_{k + 1}}] - \mathbb{E}[{V_k}] \\
&=  - J({\theta _{k + 1}}) + \mathbb{E}\left\| {{{\omega }_{k + 1}} - \omega _{k + 1}^*} \right\|_2^2 + J({\theta _k}) - \mathbb{E}\left\| {{{\omega }_k} - \omega _k^*} \right\|_2^2\\
& \le  - \frac{\alpha_k }{2}\left\| {\nabla J({\theta _k})} \right\|_2^2 + \frac{{20{\alpha_k }{H^2}\left( {1 + (\eta  - 1)\rho } \right)}}{{M(1 - \rho )}} + 10{\alpha_k }\left( {C_g^2 + 2\left( {{\chi ^2} + 1} \right)} \right)\left\| {{{\omega }_{k + 1}} - \omega _{k + 1}^*} \right\|_2^2 + \frac{{20{\alpha_k }{\kappa ^2}}}{{{c^2}}}+ \frac{5{\alpha_k }}{2}{\xi _{critic}} \\
& - \frac{\alpha_k }{2}(1-{\alpha_k}L-20\alpha _k^2L_{\omega}^2)\left\| {\frac{1}{N}\sum\limits_{i = 1}^N {{{\hat h}^i}({{\omega }_{k+1}},{\theta_k})} } \right\|_2^2 + \mathbb{E}\left\| {{{\omega }_{k + 1}} - \omega _{k + 1}^*} \right\|_2^2 - \mathbb{E}\left\| {{{\omega }_k} - \omega _k^*} \right\|_2^2
\end{align*}
Applying the error of lower level, we have
\begin{align*}
&\mathbb{E}[{V_{k + 1}}] - \mathbb{E}[{V_k}]\\
 \le & - \frac{\alpha_k }{2}\left\| {\nabla J({\theta _k})} \right\|_2^2 + \frac{{20{\alpha_k }{H^2}\left( {1 + (\eta  - 1)\rho } \right)}}{{M(1 - \rho )}} + \frac{{20{\alpha_k }{\kappa ^2}}}{{{c^2}}} + \left[ {C(\chi)C({\alpha _k}) {\left( {1 - \frac{{\beta_k \bar \upsilon \lambda }}{4}} \right)^T} - 1} \right]\left\| {{{\omega }_k} - \omega _k^*} \right\|_2^2\\
& - \frac{{{\alpha _k}}}{2}\left( {\frac{1}{2} - {\alpha _k}L - 20\alpha _k^2L_\omega ^2 - \frac{{{L_{\omega \theta }}{\alpha _k}}}{2} - 2{\alpha _k}L_\omega ^2} \right)\left\| {\frac{1}{N}\sum\limits_{i = 1}^N {{{\hat h}^i}({{\omega }_{k+1}},{\theta_k})} } \right\|_2^2+\frac{5{\alpha_k }}{2}{\xi _{critic}}\\
&+C(\chi)C({\alpha _k})T\left( {{C_1}{\beta_k ^4} + {C_2}{\beta_k ^3} + {C_3}\frac{{{\beta_k ^2}}}{N} + {C_4}{\beta_k ^2}} \right)
\end{align*}
where $C({\alpha _k}) = 1+ {\alpha_k}L_\omega^2 + \frac{{ {L_{\omega\theta}}}\alpha _k^2C_f^2}{4}$ and $C(\chi) = {20\left( {{\chi ^2} + 1} \right) + 10C_g^2}$.

If we select 
${\alpha _k} = \min \left\{ {\frac{1}{{{L_{\omega \theta }} + 2L + 4L_\omega ^2 + 40L_\omega ^2}},\alpha \sqrt {\frac{N}{K}} } \right\}$, ${\beta _k} = \frac{{4L_\omega ^2 + {L_{\omega \theta }}C_f^2}}{{T\bar v\lambda }}{\alpha _k}$
which ensures that
\begin{align*}
{\frac{1}{2} - {\alpha _k}L - 20\alpha _k^2L_\omega ^2 - \frac{{{L_{\omega \theta }}{\alpha _k}}}{2} - 2{\alpha _k}L_\omega ^2}\ge0 \\
\frac{{T{\beta _k}\bar v\lambda }}{4} \ge {\alpha_k}L_\omega^2 + \frac{{ {L_{\omega\theta}}}\alpha _k^2C_f^2}{4}
\end{align*}
We can simplify the last inequality as
\begin{align*}
&\mathbb{E}[{V_{k + 1}}] - \mathbb{E}[{V_k}]\\
&  \le  - \frac{{{\alpha _k}}}{2}\left\| {\nabla J({\theta _k})} \right\|_2^2 + \frac{{20{\alpha _k}{H^2}\left( {1 + (\eta  - 1)\rho } \right)}}{{M(1 - \rho )}} + \frac{{20{\alpha _k}{\kappa ^2}}}{{{c^2}}} + \frac{{5{\alpha _k}{\xi _{critic}}}}{2}\\
&+ TC(\chi)C({\alpha _k})\left( {{C_1}{\beta_k ^4} + {C_2}{\beta_k ^3} + {C_3}\frac{{{\beta_k ^2}}}{N} + {C_4}{\beta_k ^2}} \right)
\end{align*}
After telescoping, we have
\begin{align*}
&\frac{1}{K}\sum\limits_{k = 1}^K \mathbb{E}\left\| {\nabla J({\theta _k})} \right\|_2^2\\
&\le \frac{{2{V_1}}}{{{\alpha _k}K}} + \frac{{40{H^2}\left( {1 + (\eta  - 1)\rho } \right)}}{{M(1 - \rho )}} + \frac{{40{\kappa ^2}}}{{{c^2}}} + {5{\xi _{critic}}} +2TC(\chi )C({\alpha _k})\left( {{C_1}\beta _k^2 + {C_2}{\beta _k} + \frac{{{C_3}}}{N} + {C_4}} \right)\frac{{\beta _k^2}}{{{\alpha _k}}}
\end{align*}
which, together with ${\alpha _k} = O\left( {\sqrt {\frac{N}{K}} } \right)$, ${\beta _k} = O\left( {\sqrt {\frac{N}{K}} } \right)$, completes the proof.

\end{document}